%% file: main.tex
\documentclass{article}
\PassOptionsToPackage{numbers,sort&compress}{natbib}
\PassOptionsToPackage{table}{xcolor}

\usepackage[main,final]{neurips_2026}

\usepackage[utf8]{inputenc} 
\usepackage[T1]{fontenc}    
\usepackage{hyperref}       
\hypersetup{hidelinks}
\usepackage{url}            
\usepackage{booktabs}       
\usepackage{amsfonts}       
\usepackage{nicefrac}       
\usepackage{microtype}      
\usepackage{xcolor}         
\usepackage{multirow}
\usepackage{graphicx}
\usepackage{bbding}          

\usepackage{booktabs}
\usepackage{multirow}
\usepackage[table]{xcolor}
\usepackage{makecell}
\usepackage{siunitx}
\usepackage{diagbox}
\usepackage{algorithm}
\usepackage{algpseudocode}
\usepackage{longtable}
\usepackage{xltabular}
\usepackage{array}
\usepackage{caption}

\newcolumntype{Y}{>{\centering\arraybackslash}X}

\usepackage{array}
\usepackage{makecell}
\usepackage{multirow}
\usepackage{calc}
\usepackage{amsmath}

\newcommand{\oursrow}{\rowcolor{blue!6}}
\newcommand{\best}[1]{\textbf{#1}}
\newcommand{\gain}[1]{\textcolor{red}{+#1}}

\title{PACE: Phase-Aware Chunk Execution for Robot Policies with Action Chunking}

\author{%
  \textbf{Junnan Nie$^{1,2}$, Jiayi Li$^{2}$, Chenghao Liu$^{1}$, Junyi Lao$^{1}$,}\\
  \textbf{Jiachen Zhang$^{1}$, Tianle Zhang$^{2}$, Songfang Huang$^{1}$\textsuperscript{\Envelope}}\\
  $^1$Peking University\qquad $^2$JD Explore Academy
}

\begin{document}

\maketitle

\begingroup
\renewcommand{\thefootnote}{\Envelope}
\footnotetext[1]{Corresponding author: Songfang Huang (\texttt{hsf@pku.edu.cn}).}
\endgroup

\begin{abstract}
Action-chunking robot policies predict sequences of future actions, but deployment requires choosing how many actions to execute before replanning. The execution horizon trades off feedback frequency against motion continuity. However, success varies non-monotonically with the execution horizon, and the best-performing horizon differs across tasks. These effects make a single fixed horizon unreliable as a default execution strategy across tasks. We propose PACE (Phase-Aware Chunk Execution), a training-free method that selects execution horizons online from predicted action chunks. PACE uses prominent low-speed valleys in the predicted speed profile as candidate replanning boundaries, exploiting the phase-dependent kinematics of manipulation trajectories. By default, PACE calibrates the valley threshold from each task's existing training demonstrations, without task-specific fixed-horizon tuning. On the 50-task RoboTwin2.0 benchmark, PACE improves average success from 57.8\% to 64.2\% over the strongest fixed-horizon baseline and outperforms the adaptive execution baselines AAC and AutoHorizon. With GR00T N1.7 on LIBERO, PACE achieves the highest overall success rate among the evaluated strategies, supporting its applicability across base policies, benchmarks, and action representations. On real robots, PACE improves the average task score from 60.7 to 77.7 and success rate from 50.7\% to 70.4\% over full-chunk execution.
Code is available on our \href{https://niejunnan25.github.io/pace-project/}{\textcolor{blue}{project page}}.
\end{abstract}

\section{Introduction}
\label{sec:intro}

Recent progress in vision-language-action models and diffusion-based
robot policies has largely come from training-side gains. Larger
backbones, broader demonstration data, and stronger multi-modal
pretraining have all improved predictive performance~\cite{brohan2023rt2,kim2024openvla,black2024pi0,chi2023diffusion,zhang2026joyaira01,joyairateam2026joyaira05,zhang2026robogene}. At
deployment, however, performance depends not only on how accurately a
policy predicts, but also on how those predictions are executed. To
improve local motion continuity, these policies often adopt action
chunking~\cite{zhao2023learning,chi2023diffusion}, where each query predicts a short sequence of future
actions rather than a single-step command.

Yet an action chunk only specifies what the policy predicts, not how much of that prediction the robot should execute.
Its length fixes the prediction horizon, but a separate choice must still be made about how many actions to execute before acquiring a new observation and querying the policy again.
We refer to this executed prefix length as the \emph{execution horizon}.
Unlike the prediction horizon, the execution horizon is neither learned nor prescribed by the policy~\cite{chi2023diffusion,zhao2023learning}. It is an inference-time choice that trades off feedback frequency against within-chunk motion continuity.
As such, the execution horizon is not a mere implementation detail, but a central variable for reliable deployment of action-chunking robot policies.

In practice, the execution horizon is typically fixed to a constant $H$ throughout an episode~\cite{chi2023diffusion,zhao2023learning}.
At each replanning step, the robot executes the first $H$ actions of the predicted chunk before querying the policy again, tying all chunks to the same replanning interval regardless of the motion they predict.
A smaller $H$ incorporates new observations more often but can interrupt a coherent local motion, while a larger $H$ follows the predicted chunk longer but delays reactive correction.
Without a principled selection criterion, $H$ is usually chosen empirically.

\begingroup
\begin{figure}[!tbp]
    \centering
    \includegraphics[width=\linewidth]{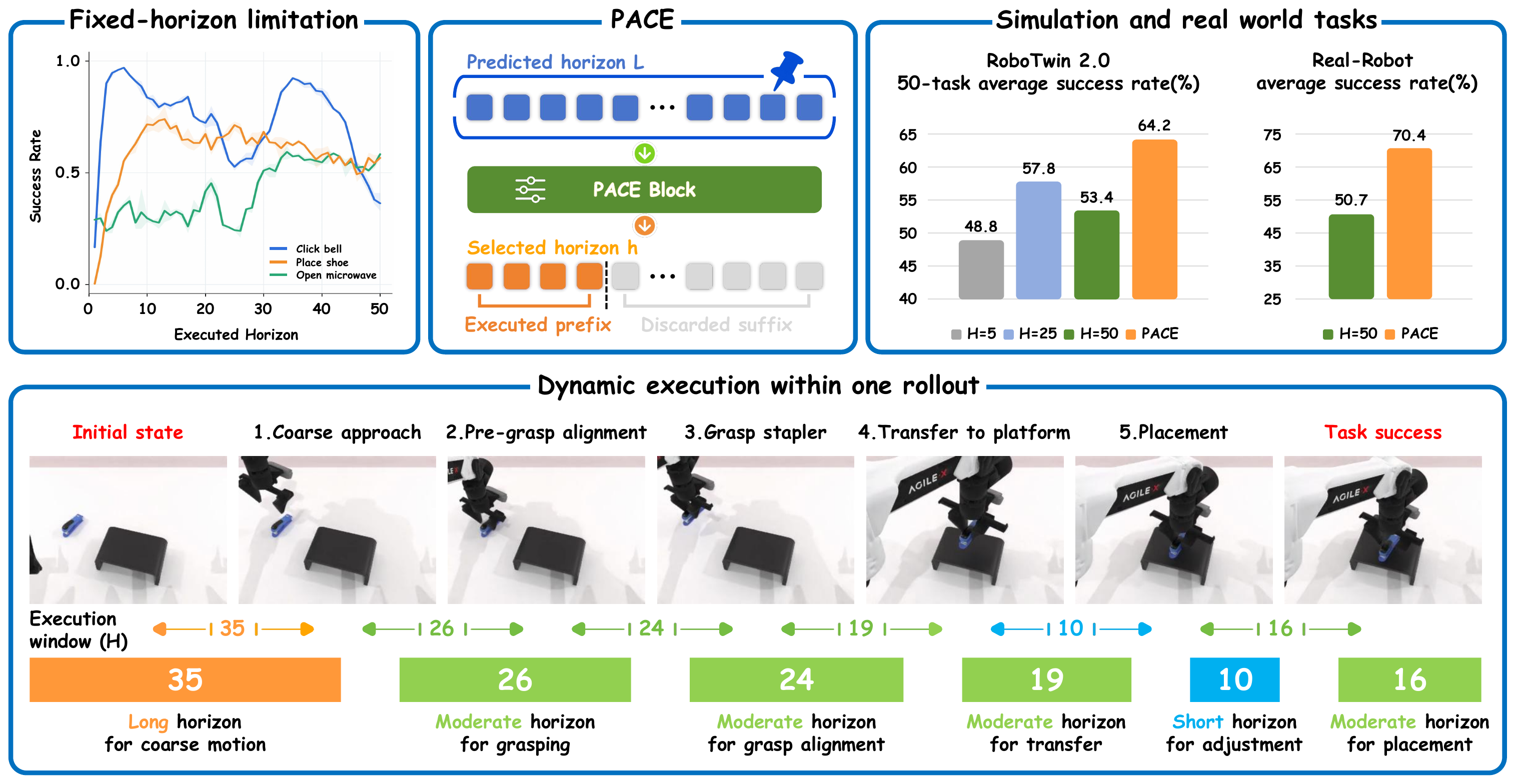}
\caption{
\textbf{Overview of PACE.}
Fixed horizons can be unreliable because success varies with \(H\) across tasks.
PACE selects the executed prefix online from each predicted chunk, improving performance in simulation and real-robot experiments while adapting the horizon in a rollout.
}
    \label{fig:intro_overview}
\end{figure}

Sweeping $H$ from $1$ to $50$ on three RoboTwin2.0 tasks reveals strongly non-monotonic success rates and task-dependent preferred horizons (Fig.~\ref{fig:intro_overview}, left). On Click Bell, success peaks near $H\approx 6$, drops by nearly 40 points around $H\approx 25$, and rebounds to a second peak near $H\approx 35$. A fixed horizon can therefore cause substantial performance degradation on some tasks. The best horizon found offline on one task also does not transfer to new tasks or real-robot deployment. These findings indicate that the execution horizon must be selected online rather than fixed in advance.

We propose PACE (Phase-Aware Chunk Execution), a test-time method that selects the execution horizon online from the speed profile of each predicted action chunk. Manipulation trajectories consist of locally coherent motion segments, with transitions often accompanied by deceleration. Policies trained via imitation learning preserve this structure in their predicted action chunks, where phase transitions can appear as low-speed valleys in the speed profile. PACE treats such valleys as candidate replanning points and executes the chunk up to the earliest accepted valley, aiming to preserve coherent motion within a segment and re-query the policy before the next begins. PACE uses no model-internal signals or learned components and requires no policy changes or retraining.

On the 50-task RoboTwin2.0 benchmark~\cite{chen2025robotwin2}, PACE improves the average success rate of \(\pi_{0.5}\) from $57.8\%$ to $64.2\%$ over the best fixed-horizon baseline and outperforms the adaptive baselines AAC and AutoHorizon.
On real robots, it improves average task score from $60.7$ to $77.7$ and success rate from $50.7\%$ to $70.4\%$ over full-chunk execution.
With GR00T N1.7 on LIBERO, PACE achieves the highest overall success rate among the evaluated strategies, supporting its applicability across policies, benchmarks, and action representations.
Ablations, sensitivity analyses, and fixed-horizon sweeps on representative tasks assess its design choices, while analyses of successful and failed rollouts reveal its working mechanism and failure modes. Our contributions are summarized as follows:

\begin{itemize}
\item We identify the execution horizon as a critical yet overlooked test-time variable, and show empirically that its effect on success is both task-dependent and non-monotonic, indicating that no single fixed horizon provides a reliable 
default across tasks.

\item We propose PACE, a plug-and-play test-time method that selects the executed prefix online from the phase-dependent kinematic structure of predicted action chunks. It requires no model-internal signals, no learned module, and no modification or retraining of the policy.

\item We validate PACE with \(\pi_{0.5}\) on all 50 RoboTwin2.0 tasks, with GR00T N1.7 on LIBERO, and on real robots. Comparisons with fixed-horizon and adaptive baselines demonstrate its effectiveness, while ablations and rollout analyses explain its performance gains.
\end{itemize}

\endgroup

\begingroup
\begin{figure}[!tbp]
\centering
\includegraphics[width=\linewidth]{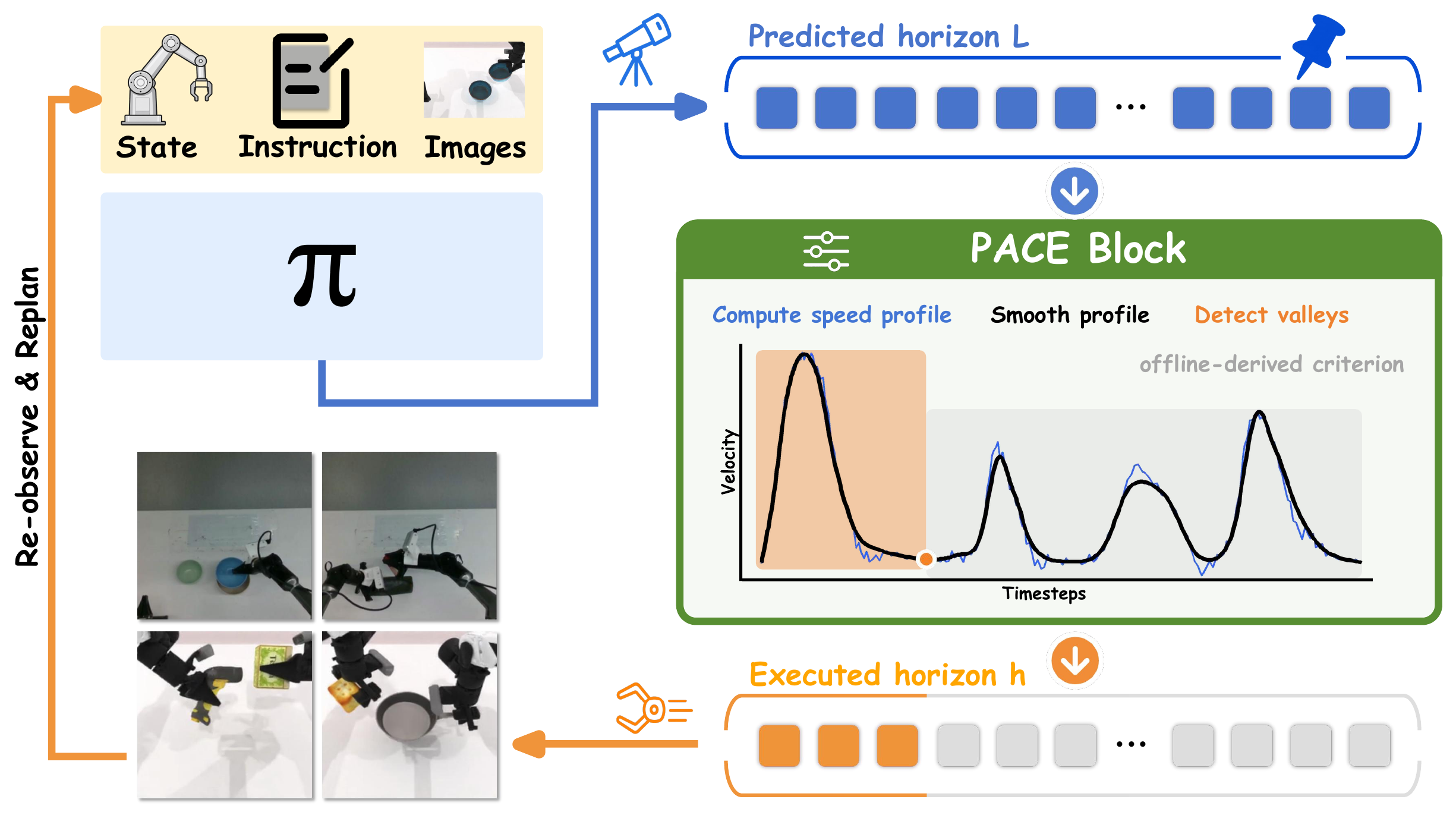}
\caption{
\textbf{PACE framework.}
PACE selects an execution horizon from low-speed valleys in the predicted chunk's smoothed speed profile. The robot executes the selected prefix, discards the suffix, and then queries the policy again.
}
\label{fig:overview}
\end{figure}

\section{Related work}
\label{sec:related}

\paragraph{Robot policies with action chunking.}
Action chunking has become a common interface in robot policy learning.
It was popularized by ACT for fine-grained bimanual imitation learning~\cite{zhao2023learning} and Diffusion Policy for receding-horizon action-sequence generation~\cite{chi2023diffusion}.
Subsequent work applies this interface to mobile manipulation~\cite{fu2024mobile}, 3D visuomotor control~\cite{ze2024dp3}, and large-scale bimanual policy learning~\cite{liu2024rdt}.
Generalist policies such as Octo, $\pi_0$, and $\pi_{0.5}$ extend chunk prediction to diverse tasks and environments~\cite{ghosh2024octo,black2024pi0,black2025pi05}.
VLA research also explores efficient action tokenization~\cite{pertsch2025fast} and faster inference for action prediction~\cite{kim2025oft,shukor2025smolvla}.
While these works focus on generating, representing, or scaling action chunks, PACE addresses the complementary deployment-time question of how much of each predicted chunk to execute before incorporating new observations and replanning.

\paragraph{Test-time execution of action chunks.}
Many policies execute action chunks using fixed test-time rules: ACT smooths overlapping predictions through temporal ensembling, while Diffusion Policy executes a fixed action horizon in a receding-horizon loop~\cite{zhao2023learning,chi2023diffusion}.
Recent inference-time methods reduce latency-induced discontinuities or stabilize chunk execution~\cite{black2025realtime,wang2026remac}.
Another line modifies training or model design to reason over multiple horizons~\cite{jing2025mixture}.
More closely related are methods that adapt the execution horizon at inference time.
Adaptive Action Chunking (AAC)~\cite{liang2026adaptive} estimates prefix-wise action entropy from multiple sampled chunks and selects the horizon subject to a minimum-movement constraint.
AutoHorizon~\cite{wang2026vla} estimates the execution horizon from action-token self-attention in flow-based VLAs, treating attention as a proxy for the model's predictive limit.
PACE instead selects replanning boundaries from prominent low-speed valleys in the kinematic profile of a single decoded action chunk, without requiring model-internal signals or additional chunk samples.

\section{Method}
\label{sec:method}

\subsection{Problem formulation}
\label{sec:problem_formulation}

We consider a manipulation policy with action chunking.
At the \(i\)-th policy query, the policy receives the observation \(o_{\tau_i}\) and the language instruction \(\ell\), and outputs
\begin{equation}
\mathbf{A}_i = \pi_\theta(o_{\tau_i}, \ell) \in \mathbb{R}^{L \times d_a},
\qquad
\mathbf{A}_i = (a_{i,1}, \ldots, a_{i,L}),
\end{equation}
where \(L\) is the prediction horizon and \(d_a\) is the action dimension.
Let \(h_i\in\{1,\ldots,L\}\) denote the execution horizon, i.e., the number of predicted actions executed before the next policy query.
After executing the prefix \((a_{i,1},\ldots,a_{i,h_i})\), the next query occurs at \(\tau_{i+1}=\tau_i+h_i\).

A fixed-horizon rule sets \(h_i^{\mathrm{fixed}}=H\) for all \(i\), whether the predicted chunk describes a smooth transit or a contact event.
PACE instead conditions the execution horizon on the predicted chunk,
\begin{equation}
h_i = g(\mathbf{A}_i), \qquad g:\mathbb{R}^{L\times d_a}\to\{1,\ldots,L\}.
\end{equation}
The deterministic, training-free function \(g\) uses fixed hyperparameters (Secs.~\ref{sec:kinematic_profile}--\ref{sec:phase_aware_selection}) without modifying the trained policy \(\pi_\theta\).
The prediction horizon \(L\) remains the policy's training horizon (Sec.~\ref{sec:ablation}).

\endgroup

\subsection{Kinematic profile extraction}
\label{sec:kinematic_profile}

Expert manipulation trajectories are rarely uniform in speed. They organize into alternating fast and slow segments, where low-speed regions typically arise under contact, stabilization, or sub-stage transitions. An imitation-trained policy tends to preserve this structure in its predicted action chunks. This makes the speed profile a kinematic cue for where one locally coherent motion segment ends and the next begins, and thus a natural basis for the selection rule \(g\).

Concretely, for each controllable arm group \(b=1,\ldots,B\), let \(q_{i,j}^{b}\in\mathbb{R}^{d_q}\) denote the joint command at predicted step \(j\). We define the speed profile \(\mathbf{v}_i^b = (v_{i,1}^b,\ldots,v_{i,L-1}^b)\) by
\begin{equation}
v_{i,j}^{b} = f_s \, \| q_{i,j+1}^{b} - q_{i,j}^{b} \|_2, \qquad j=1,\ldots,L-1,
\end{equation}
where \(f_s\) is the control frequency. Raw speed profiles contain high-frequency fluctuations that create spurious local valleys and, if untreated, trigger overly frequent replanning. We therefore apply a Savitzky--Golay filter \(\mathcal{S}\) to obtain the smoothed profile \(\tilde{\mathbf{v}}_i^b = \mathcal{S}(\mathbf{v}_i^b)\), whose low-speed valleys serve as candidate replanning boundaries for the selection rule in Sec.~\ref{sec:phase_aware_selection}. We use joint-space speed by default; a Cartesian variant yields comparable success after smoothing (Sec.~\ref{sec:ablation}), and filter parameters are given in Appendix~\ref{app:adaptive_horizon_details}.

\subsection{Phase-aware execution horizon selection}
\label{sec:phase_aware_selection}

Given the kinematic profile from Sec.~\ref{sec:kinematic_profile}, we now ask where inside the chunk the policy should stop and replan.
Our guiding observation, illustrated in Fig.~\ref{fig:overview}, is that phase boundaries such as contacts, grasps, and releases can manifest as pronounced low-speed valleys in \(\tilde{\mathbf{v}}_i^b\): the end-effector decelerates as it approaches the event and accelerates again once it is resolved.
The function \(g\) therefore searches for the earliest such valley and treats it as a candidate point to hand control back to the policy.

Not every local minimum qualifies: a valley is informative only when preceded by a clearly faster segment, otherwise it merely reflects noise on an already slow trajectory.
Let \(\mathcal{V}_i^b\subseteq\{1,\ldots,L-1\}\) denote the local minima of \(\tilde{\mathbf{v}}_i^b\) satisfying a minimum spacing constraint.
For each candidate \(r\in\mathcal{V}_i^b\) we define the preceding maximum
\begin{equation}
m_i^b(r) = \max_{1\le j\le r} \tilde v_{i,j}^b,
\end{equation}
and the contrast score
\begin{equation}
\Delta_i^b(r)
= \min\!\bigl\{\,m_i^b(r)-\tilde v_{i,1}^b,\; m_i^b(r)-\tilde v_{i,r}^b\,\bigr\},
\end{equation}
which requires the speed to both rise substantially before the valley and drop substantially into it, so that \(r\) sits at the bottom of a genuine deceleration rather than on a flat region.
A valley is accepted when \(\Delta_i^b(r)\ge\delta_{\mathcal{T}}\), where \(\delta_{\mathcal{T}}\) is calibrated once per task from offline training trajectories using a fixed percentile rule shared across tasks.

Accepted valleys from all arm groups are pooled into
\begin{equation}
\mathcal{R}_i
= \bigcup_{b=1}^{B}
\bigl\{\, r\in\mathcal{V}_i^b \,:\, \Delta_i^b(r)\ge \delta_{\mathcal{T}} \,\bigr\}.
\end{equation}
We take the union because a candidate phase transition on any single arm is already a reason to replan, and we select the earliest candidate, aiming to query the policy before, rather than after, a boundary is crossed with stale observations.
When no valley is accepted, PACE falls back to the maximum execution horizon, giving
\begin{equation}
h_i = g(\mathbf{A}_i) =
\begin{cases}
\min \mathcal{R}_i, & \mathcal{R}_i \ne \emptyset,\\
H_{\max}, & \mathcal{R}_i = \emptyset,
\end{cases}
\end{equation}
where \(H_{\max}\le L\) is the execution horizon used when no valley is accepted.
Calibration of \(\delta_{\mathcal{T}}\), the valley-spacing constant, and the value of \(H_{\max}\) are deferred to Appendix~\ref{app:adaptive_horizon_details}.

\section{Simulation experiments}
\label{sec:sim_experiments}

\paragraph{Experimental setup.}
We evaluate PACE on 50 RoboTwin2.0 bimanual manipulation tasks, training one \(\pi_{0.5}\) checkpoint per task under the standard protocol.
All methods share the same checkpoint and differ only in test-time execution.
Unless otherwise noted, our \emph{full evaluation protocol} uses 900 episodes per task (three seeds, each with three batches of 100 episodes), with success rates averaged over all 50 tasks.
Task-level tables show six representative tasks balanced across short-, medium-, and long-duration groups.
Full results and training details are in Appendix~\ref{app:sim_full_results}.

\paragraph{Overview of experiments.}
We compare PACE as a default execution strategy with fixed-horizon and adaptive baselines (Sec.~\ref{sec:main_results}) and analyze its horizon adaptation within rollouts (Sec.~\ref{sec:dynamic_window_visualization}).
We ablate the phase profile, training prediction horizon, and threshold sharing (Sec.~\ref{sec:ablation}), and use task-specific fixed-horizon sweeps as a diagnostic upper bound (Sec.~\ref{sec:fixed_horizon_sweep}).

\begin{table*}[!t]
\centering
\small
\setlength{\tabcolsep}{3.435376pt}
\renewcommand{\arraystretch}{1.15}
\setlength{\heavyrulewidth}{0.65pt}
\setlength{\lightrulewidth}{0.3pt}
\setlength{\cmidrulewidth}{0.3pt}
\caption{\textbf{RoboTwin2.0 results with $\pi_{0.5}$.} Success rates (\%) on six representative tasks, grouped by task duration, and all 50 tasks. Overall is the mean success rate over all 50 tasks. Bold marks the highest mean in each column. Red values after the slash report PACE improvements over the best of $\mathrm{H}=5,25,50$ in each column (percentage points, computed before rounding).}
\label{tab:main_results}
\begin{tabular}{lccccccc}
\toprule
\multirow{2}{*}{Method} & \multicolumn{2}{c}{Short} & \multicolumn{2}{c}{Medium} & \multicolumn{2}{c}{Long} & \multirow{2}{*}{\makecell{Overall\\(50 tasks)}} \\
\cmidrule(lr){2-3}\cmidrule(lr){4-5}\cmidrule(lr){6-7}
& Pick Bottles & Grab Roller & Place A2B & Dump Bin & Put Bottles & Stack Bowls & \\
\midrule
\makebox[54pt][l]{$\mathrm{H}=5$} & 13.3 & 89.9 & 20.0 & 58.7 & 0.3 & 88.7 & 48.8 \\
\makebox[54pt][l]{$\mathrm{H}=25$} & 45.3 & 96.2 & 47.8 & 85.1 & 79.3 & 88.4 & 57.8 \\
\makebox[54pt][l]{$\mathrm{H}=50$} & 38.7 & 96.2 & 36.8 & 84.3 & 70.8 & 85.1 & 53.4 \\
\midrule
\makebox[54pt][l]{AutoHorizon} & 41.6 & 96.9 & 43.1 & 88.1 & 74.3 & 91.0 & 59.0 \\
\makebox[54pt][l]{AAC} & \textbf{48.9} & 97.7 & 46.7 & 89.1 & 76.6 & 88.2 & 59.1 \\
\rowcolor{blue!6}\makebox[54pt][l]{\textbf{PACE} (Ours)} & 48.2 / \textcolor{red}{+2.9} & \textbf{98.6} / \textcolor{red}{+2.3} & \textbf{54.6} / \textcolor{red}{+6.8} & \textbf{93.0} / \textcolor{red}{+7.9} & \textbf{88.9} / \textcolor{red}{+9.6} & \textbf{95.6} / \textcolor{red}{+6.9} & \textbf{64.2} / \textcolor{red}{+6.4} \\
\bottomrule\end{tabular}
\end{table*}

\subsection{Simulation results}
\label{sec:main_results}

We compare PACE against fixed-horizon execution with \(H\in\{5,25,50\}\), corresponding to short, intermediate, and full-chunk prefixes (recall \(H_{\text{train}}{=}50\)).
We also evaluate Adaptive Action Chunking (AAC)~\cite{liang2026adaptive} and AutoHorizon~\cite{wang2026vla} using the same task-specific \(\pi_{0.5}\) checkpoints.
Our bimanual joint-space adaptation of AAC estimates action entropy from 20 sampled chunks per policy query.
PACE calibrates one threshold per task from the checkpoint's existing training demonstrations, without using evaluation rollouts, success labels, or a fixed-horizon search (Appendix~\ref{app:threshold_calibration}).

Table~\ref{tab:main_results} shows that PACE achieves the highest 50-task average, exceeding the best global fixed horizon (\(H=25\)) by 6.4 points and AAC and AutoHorizon by approximately 5.1 and 5.2 points, respectively.
PACE also exceeds the strongest fixed horizon on every displayed task and achieves the highest mean on five of the six.
AAC leads on Pick Bottles (48.9\% versus 48.2\%).

\paragraph{Cross-policy evaluation.}
We evaluate PACE with GR00T N1.7~\citep{nvidia2026grootn17} on all four LIBERO suites~\citep{liu2023libero}, covering 40 tasks.
GR00T predicts 16-step Cartesian-delta chunks.
Without retraining or modifying GR00T, PACE constructs its motion profile from the translational-delta norm and applies the same valley-based selection rule.
PACE achieves the highest overall success rate, \(95.33\%\), compared with \(94.47\%\) for the strongest global fixed horizon (\(H=16\)) and \(94.92\%\) for AutoHorizon.
These results support the applicability of PACE across different base policies, benchmarks, action representations, and chunk lengths.
The horizon selector takes \(0.0566\) ms per query, approximately \(0.044\%\) of the time required for a single policy forward pass (Appendix~\ref{app:libero}).

\begingroup
\begin{figure}[!t]
\centering
\includegraphics[width=\linewidth]{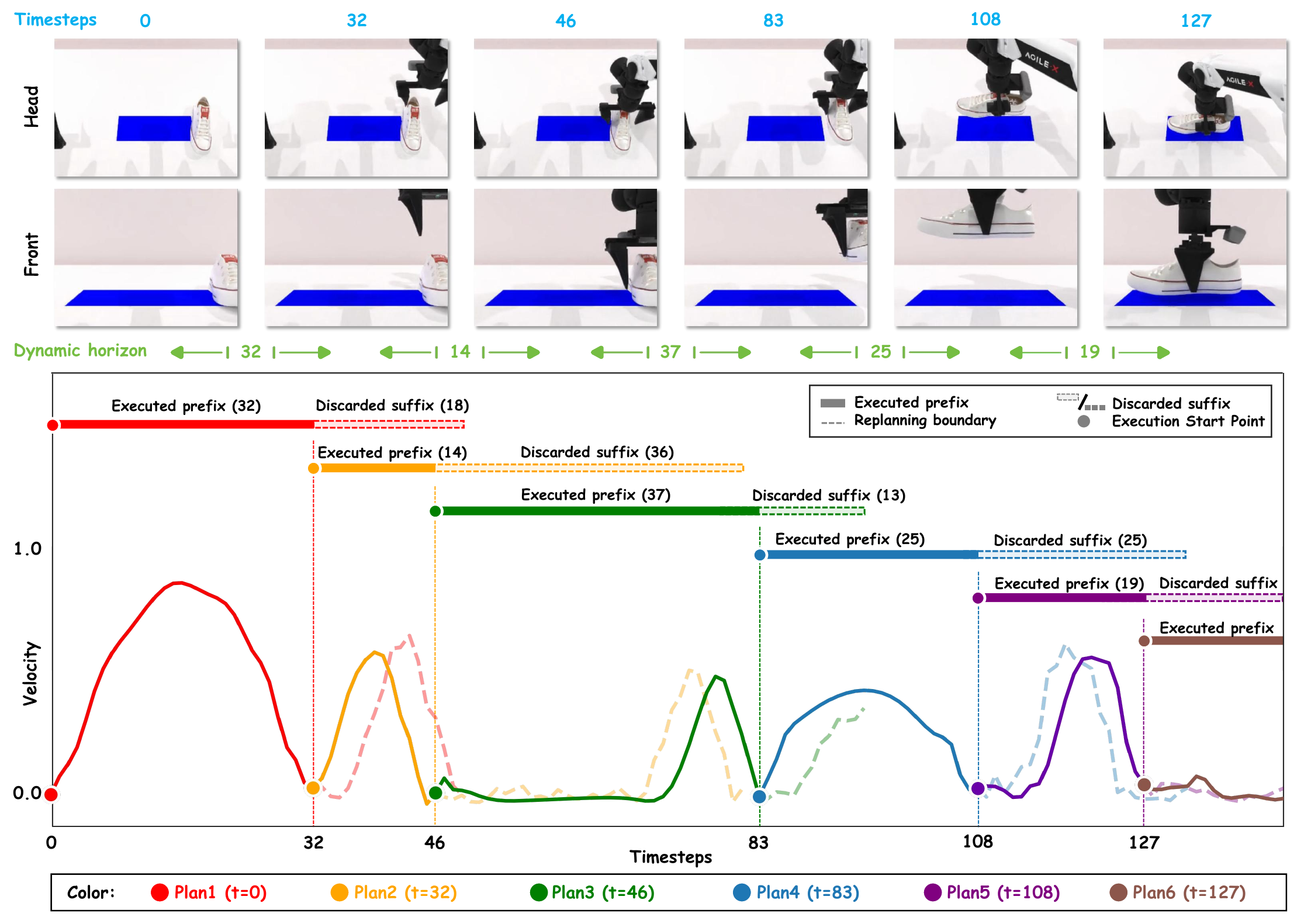}
\caption{
\textbf{Rollout-level behavior of PACE.}
The rollout is from \texttt{place\_shoe}.
The top rows show head and front camera observations at the six replanning timesteps ($t=0, 32, 46, 83, 108, 127$), with the selected execution horizon annotated between consecutive queries.
The bottom panel shows the six predicted action chunks along the rollout timeline, one color per plan.
For each plan, the horizontal bar marks the executed prefix (solid) and discarded suffix (dashed), and the curve below shows the corresponding predicted velocity profile under the same solid/dashed convention.
Circles mark the execution start point of each plan, and vertical dashed lines mark the replanning boundaries.
}
\label{fig:adaptive_rollout}
\end{figure}

\subsection{Visualizing adaptive execution horizons}
\label{sec:dynamic_window_visualization}

We visualize a representative \texttt{place\_shoe} rollout in Fig.~\ref{fig:adaptive_rollout} to examine how PACE adapts its execution horizon.
Selected horizons range from 14 to 37 steps, with longer horizons covering coherent motion (the approach in Plan\,3, 37 steps) and shorter horizons near low-speed transitions (Plan\,2, 14 steps).
A fixed horizon cannot adapt to this within-episode structure.

The discarded suffixes show how PACE limits open-loop execution as the scene changes.
PACE replans at boundaries selected from the predicted motion structure, and the new plan often differs from the discarded suffix over the overlapping interval, as seen in the mismatched solid and dashed curves of adjacent plans around $t{=}83$.
This limits reliance on stale observations while preserving long horizons for coherent motion, helping explain the gains over fixed-horizon baselines in Sec.~\ref{sec:main_results}.
Appendix~\ref{app:phase_alignment} quantifies the alignment of replanning boundaries with motion-phase transitions.

\subsection{Ablation analysis}
\label{sec:ablation}

We examine how motion-profile construction, the training prediction horizon, and threshold sharing affect PACE's performance.
We evaluate motion-profile variants and compare shared versus per-task thresholds on all 50 RoboTwin2.0 tasks, with unchanged policy checkpoints.
To study the training prediction horizon \(H_{\mathrm{train}}\), we train policies to predict action chunks of different lengths.
We compare their success rates at the same fixed execution horizon to assess whether supervision over longer action sequences improves the quality of the executed prefix.
Since each setting requires retraining, we evaluate them on six representative tasks.
Further details and results appear in Appendix~\ref{app:ablation_details}.

\endgroup

\begingroup
\begin{table}[!t]
\centering
\small
\setlength{\tabcolsep}{3.4pt}
\renewcommand{\arraystretch}{1.12}
\caption{
\textbf{Ablation studies.}
\textbf{Left:} phase-profile ablation on all 50 RoboTwin2.0 tasks; Avg.\ Horizon is the mean selected execution horizon.
\textbf{Right:} training-horizon ablation on six representative tasks; each row fixes \(H_{\mathrm{eval}}\) and varies \(H_{\mathrm{train}}\).
Gain@50 is the gain of \(H_{\mathrm{train}}{=}50\) over the shortest feasible setting.
Success rates are in \%. \(^\dagger\) marks the default.
}
\label{tab:design_analysis}
\begin{minipage}[t]{0.39\linewidth}
\vspace{0pt}
\centering
\setlength{\tabcolsep}{3.0pt}
\begin{tabular}{@{}lcc@{}}
\toprule
Configuration & Succ. & Avg. Horizon \\
\midrule
Cartesian, raw       & 62.2 & 11.3 \\
Cartesian, smoothed  & 64.5 & 19.5 \\
Joint, raw           & 60.4 & 16.7 \\
\oursrow Joint, smoothed$^\dagger$ & 64.2 & 24.3 \\
\bottomrule
\end{tabular}
\end{minipage}
\hfill
\begin{minipage}[t]{0.57\linewidth}
\vspace{0pt}
\centering
\setlength{\tabcolsep}{3.6pt}
\begin{tabular}{@{}ccccccc@{}}
\toprule
\(H_{\mathrm{eval}}\backslash H_{\mathrm{train}}\)
& 10 & 20 & 30 & 40 & 50 & Gain@50 \\
\midrule
10 & 57.2 & 68.7 & 69.9 & 74.6 & 77.9 & \textbf{+20.7} \\
20 & --   & 68.7 & 72.6 & 73.8 & 75.3 & \textbf{+6.6}  \\
30 & --   & --   & 70.2 & 71.4 & 73.8 & \textbf{+3.6}  \\
40 & --   & --   & --   & 70.1 & 73.4 & \textbf{+3.3}  \\
\bottomrule
\end{tabular}
\end{minipage}
\end{table}

\paragraph{Phase-profile construction.}
We compare joint-space and Cartesian speed profiles, with and without smoothing, to examine whether PACE depends on a specific motion representation and how smoothing affects boundary selection.
The Cartesian variant computes end-effector linear speed through forward kinematics.
We keep the valley-selection rule and calibration percentile fixed across all four variants.
Table~\ref{tab:design_analysis} (left) shows that smoothing improves success in both spaces and increases the mean selected execution horizon.
This pattern suggests that smoothing suppresses noisy local valleys that would otherwise trigger overly frequent replanning.
After smoothing, Cartesian and joint-space profiles achieve comparable success, while the joint-space variant selects longer execution horizons on average.
We adopt the smoothed joint-space profile by default.
It is within \(0.3\) points of the best variant, matches the policy action space, and avoids forward kinematics at every replanning query.

\paragraph{Training prediction horizon.}
Each row of Table~\ref{tab:design_analysis} (right) fixes \(H_{\mathrm{eval}}\)
and varies \(H_{\mathrm{train}}\); pairs with \(H_{\mathrm{train}}<H_{\mathrm{eval}}\) 
are infeasible because the policy predicts too few actions.
Success increases monotonically with \(H_{\mathrm{train}}\), 
and the gain of \(H_{\mathrm{train}}{=}50\) over the shortest feasible target 
shrinks from \(+20.7\) at \(H_{\mathrm{eval}}{=}10\) to \(+3.3\) at 
\(H_{\mathrm{eval}}{=}40\).
This trend shows that longer training targets provide stronger 
sequence-level supervision: they constrain not just the action after the observation, 
but how motion evolves across a longer expert segment. Since the executed prefix is predicted jointly 
with the rest of the chunk, it inherits this structure. 
We adopt \(H_{\mathrm{train}}{=}50\) as the default prediction horizon.

\begingroup
\begin{table}[H]
\centering
\small
\setlength{\tabcolsep}{5pt}
\renewcommand{\arraystretch}{1.15}
\caption{\textbf{Shared versus per-task threshold calibration.}
Success rates (\%), reported as mean \(\pm\) sample SD across three evaluation seeds.}
\label{tab:shared_threshold}
\begin{tabular}{@{}lll@{}}
\toprule
Execution & Threshold calibration & Success (\%) \\
\midrule
Fixed \(H=25\) & Not applicable & \(57.80 \pm 1.49\) \\
PACE, shared & Pooled training demonstrations & \(62.44 \pm 0.92\) \\
\oursrow PACE, per-task & Each task's training demonstrations & \(\mathbf{64.20} \pm 0.88\) \\
\bottomrule
\end{tabular}
\end{table}
\endgroup

\paragraph{Shared versus per-task thresholds.}
By default, PACE calibrates a separate valley-acceptance threshold from each task's training demonstrations.
To assess how much PACE depends on per-task calibration, we pool the training demonstrations from all 50 tasks to calibrate a single threshold and apply it unchanged across tasks.
All other PACE settings and evaluation conditions remain unchanged.
The shared threshold governs valley acceptance, while execution horizons still adapt to the speed profile of each predicted action chunk.
Table~\ref{tab:shared_threshold} shows that the shared threshold retains most of PACE's gain over fixed \(H=25\), improving success by \(4.64\) percentage points.
Per-task calibration provides a further gain of approximately \(1.8\) points and remains our default (Appendix~\ref{app:shared_threshold}).

\endgroup

\begingroup
\subsection{PACE versus fixed-horizon sweeps}
\label{sec:fixed_horizon_sweep}

We compare PACE with task-specific fixed-horizon sweeps on six representative tasks.
For each task, we evaluate \(H=1,\ldots,50\) to obtain the success-rate curve and the best observed fixed-horizon performance.
These results serve as diagnostic references for evaluating PACE, which selects execution horizons online without task-specific fixed-horizon search.

\begin{figure}[!t]
\centering
\includegraphics[width=0.9\linewidth]{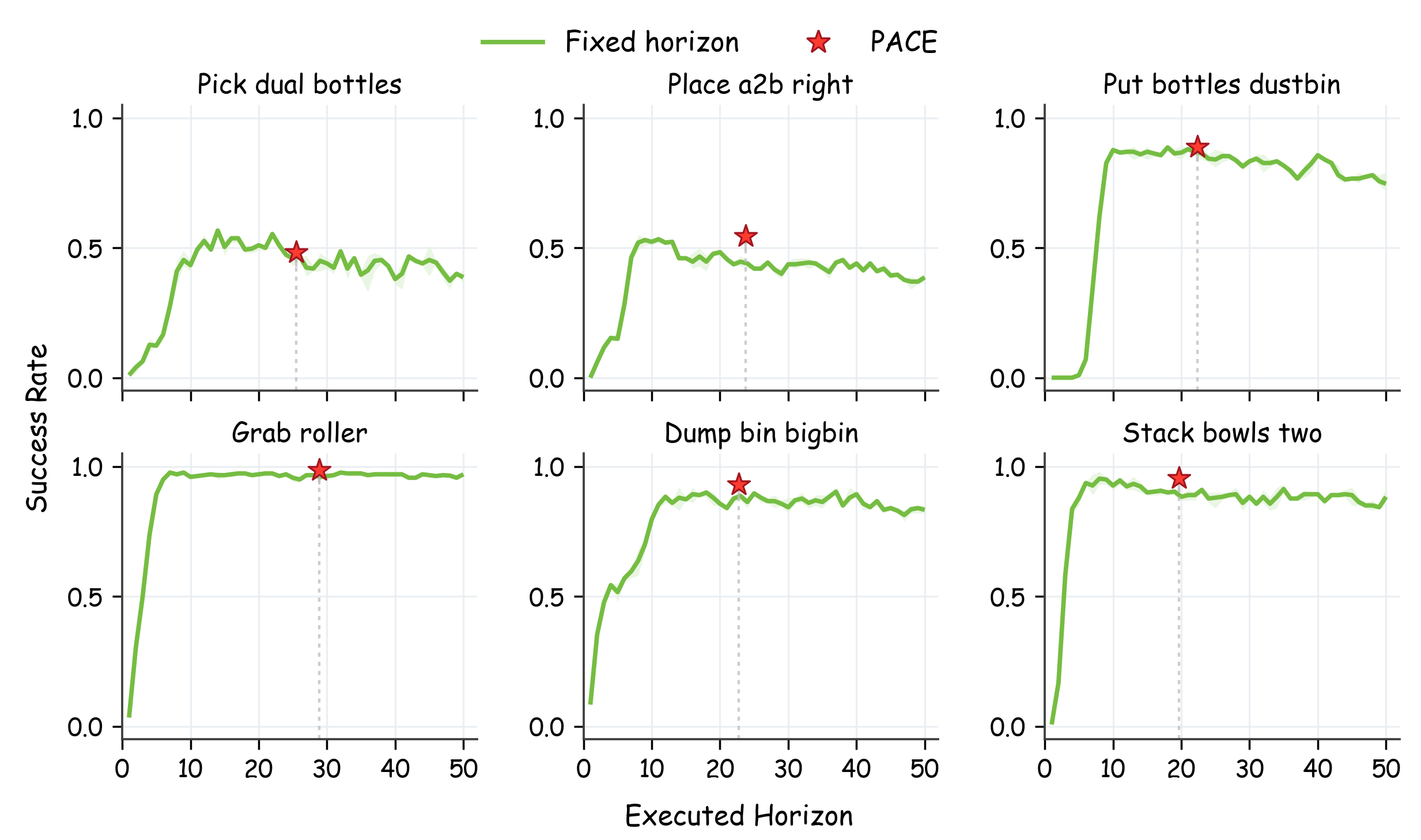}
\caption{
\textbf{PACE compared with fixed-horizon sweeps.}
Each panel shows one of the six representative tasks.
The green curve reports success as the fixed execution horizon \(H\) varies from 1 to 50.
The red star marks PACE's success rate at its mean selected execution horizon.
}
\label{fig:fixed_horizon_sweep}
\end{figure}

Figure~\ref{fig:fixed_horizon_sweep} shows that the success curves and preferred horizons vary substantially across tasks.
For example, the best fixed horizon is \(H=8\) on Stack Bowls Two but \(H=37\) on Dump Bin Bigbin.
PACE reaches the high-success region of each task without task-specific fixed-horizon tuning, with mean execution horizons ranging from \(19.7\) to \(28.9\) steps.
These results are achieved without uniformly short execution prefixes.

\begin{table}[!htbp]
\centering
\small
\setlength{\tabcolsep}{3.4pt}
\renewcommand{\arraystretch}{1.15}
\setlength{\heavyrulewidth}{0.65pt}
\setlength{\lightrulewidth}{0.3pt}
\setlength{\cmidrulewidth}{0.3pt}
\caption{\textbf{Matched-horizon comparison on RoboTwin2.0.}
For each task, the fixed horizon is obtained by rounding PACE's mean execution horizon to the nearest integer.
Red values after the slash report PACE's success-rate gains over the matched baseline.}
\label{tab:matched_horizon_main}
\begin{tabular}{lcccccc}
\toprule
\multirow{2}{*}{Setting} & \multicolumn{2}{c}{Short} & \multicolumn{2}{c}{Medium} & \multicolumn{2}{c}{Long} \\
\cmidrule(lr){2-3}\cmidrule(lr){4-5}\cmidrule(lr){6-7}
& Pick Bottles & Grab Roller & Place A2B & Dump Bin & Put Bottles & Stack Bowls \\
\midrule
Fixed $\mathrm{H}$ & 26 & 29 & 24 & 23 & 22 & 20 \\
PACE mean $\mathrm{H}$ & 25.5 & 28.9 & 23.8 & 22.7 & 22.3 & 19.7 \\
\midrule
Fixed (\%) & 46.7 & 96.0 & 44.0 & 88.7 & 87.0 & 88.3 \\
\rowcolor{blue!6}\textbf{PACE} (\%) & \textbf{48.2} / \textcolor{red}{+1.6} & \textbf{98.6} / \textcolor{red}{+2.6} & \textbf{54.6} / \textcolor{red}{+10.6} & \textbf{93.0} / \textcolor{red}{+4.3} & \textbf{88.9} / \textcolor{red}{+1.9} & \textbf{95.6} / \textcolor{red}{+7.2} \\
\bottomrule
\end{tabular}
\end{table}

\paragraph{Matched execution horizons.}
To test whether PACE's performance can be reproduced by a constant execution length, we round its mean horizon on each task to the nearest integer and evaluate that fixed horizon (Appendix~\ref{app:fixed_horizon_sweep}).
Despite similar mean horizons, the baseline replans at fixed intervals, whereas PACE selects a horizon from each predicted chunk.
Table~\ref{tab:matched_horizon_main} shows higher observed success for PACE on all six tasks.
On Place A2B Right, PACE achieves \(54.6\%\) success with a mean horizon of \(23.8\) steps, versus \(44.0\%\) for fixed \(H=24\).
These results suggest that matching the mean horizon alone is insufficient to reproduce PACE's performance.

\endgroup

\clearpage

\section{Real-robot experiments}
\label{sec:real_world}

We evaluate PACE on real robots in two settings:
RoboChallenge~\cite{yakefu2025robochallenge} and our in-lab
\texttt{place\_object\_on\_plate} task family. RoboChallenge is a
public real-robot benchmark; we evaluate \texttt{stack\_bowls} and
\texttt{put\_pen\_into\_pencil\_case} with 30 trials per method per
task. In \texttt{place\_object\_on\_plate}, a Franka robot places one of
five objects onto a plate in our laboratory setup. Within each task,
all methods use the same fine-tuned \(\pi_{0.5}\) checkpoint and differ
only in the test-time execution scheme. We report task Score and full-completion rate Succ. for both settings; task definitions, scoring rules, training details, and
per-object in-lab results appear in
Appendix~\ref{app:real_world_details}.

\subsection{Real-robot results}
\label{sec:real_world_results}

Table~\ref{tab:real_world_results} compares PACE with the full-chunk execution baseline (\(H=50\)) on the three real-robot tasks.
PACE improves both task score and success rate on every task, raising the average success rate from 50.7\% to 70.4\% and the average task score from 60.7 to 77.7, with each task weighted equally.

\begin{table}[!tbp]\centering
\small
\setlength{\tabcolsep}{3pt}
\renewcommand{\arraystretch}{1.15}
\setlength{\heavyrulewidth}{0.65pt}
\setlength{\lightrulewidth}{0.3pt}
\setlength{\cmidrulewidth}{0.3pt}
\caption{\textbf{Real-robot results.} Task scores and full-task success rates (\%) on RoboChallenge and the Franka task. Red values after the slash indicate gains over fixed $\mathrm{H}=50$.}
\label{tab:real_world_results}
\setlength{\tabcolsep}{6pt}
\begin{tabular}{lcccccc}\toprule
\multirow{2}{*}{Method} & \multicolumn{2}{c}{Stack Bowls} & \multicolumn{2}{c}{Put Pen in Case} & \multicolumn{2}{c}{Place Object on Plate} \\
\cmidrule(lr){2-3}\cmidrule(lr){4-5}\cmidrule(lr){6-7}
& Score & Succ. & Score & Succ. & Score & Succ. \\ \midrule
Baseline & 73.0 & 70.0 & 37.2 & 10.0 & 72.0 & 72.0 \\
\rowcolor{blue!6}\textbf{PACE} & \textbf{93.5} / \textcolor{red}{+20.5} & \textbf{90.0} / \textcolor{red}{+20.0} & \textbf{51.7} / \textcolor{red}{+14.5} & \textbf{33.3} / \textcolor{red}{+23.3} & \textbf{88.0} / \textcolor{red}{+16.0} & \textbf{88.0} / \textcolor{red}{+16.0} \\
\bottomrule\end{tabular}
\end{table}

To assess whether more frequent replanning with a fixed horizon yields similar performance, we further evaluate \(H=5\) and \(H=25\) on the Franka task.
PACE achieves an 88\% success rate, compared with 87\% for \(H=5\) and 82\% for \(H=25\).
It thus attains a success rate comparable to \(H=5\) while requiring 84.6\% fewer policy queries per successful episode (Appendix~\ref{app:franka_stronger_baselines}).

\subsection{PACE horizon selection in a successful rollout}
\label{sec:real_world_success_rollout}

\begin{figure}[!tbp]
\centering
\includegraphics[width=\linewidth]{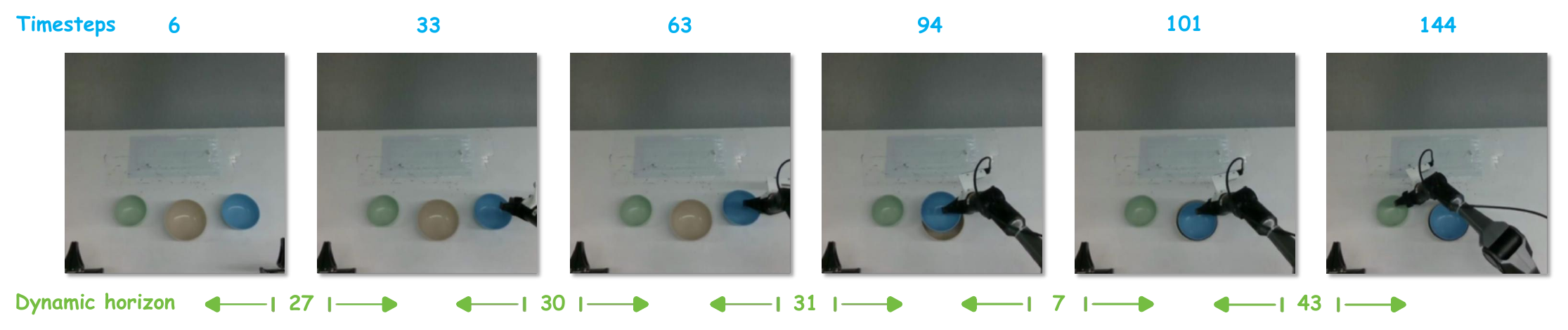}
\caption{
\textbf{Successful rollout.}
The rollout is from \texttt{stack\_bowls}. Blue: policy-query timestep.
Green: horizon length. The horizon is long during approach and transport
(27, 30, 31), short at contact-sensitive alignment (7), and long again
afterward (43).
}
\label{fig:real_world_success_rollout}
\end{figure}

Figure~\ref{fig:real_world_success_rollout} visualizes a successful 
\texttt{stack\_bowls} rollout and shows that PACE adjusts its 
execution horizon according to the local manipulation phase rather than a fixed replanning cadence. During approach and transport, the method selects 
long horizons (27, 30, 31 actions), allowing the robot to maintain continuous 
motion while open-loop execution remains reliable. Near contact-sensitive 
alignment, where small pose errors determine whether the bowl stacks, the 
horizon contracts to 7 actions so the next query can incorporate an updated 
observation. Once alignment is resolved, the horizon expands again to 43, 
showing that the method does not remain in a high-frequency replanning mode 
when a coherent motion segment becomes available.

This behavior mirrors the simulation analysis in
Sec.~\ref{sec:dynamic_window_visualization} and helps explain the real-robot
gains in Table~\ref{tab:real_world_results}: PACE introduces
feedback where stale observations are most costly, while avoiding unnecessary
queries during smooth motion. Additional rollout visualizations for representative RoboTwin2.0 and RoboChallenge tasks are provided in Appendix~\ref{app:additional_rollout_visualizations}.

\subsection{Failure case and scope of test-time execution control}
\label{sec:real_world_failure_case}

\begin{figure}[!tbp]
\centering
\includegraphics[width=\linewidth]{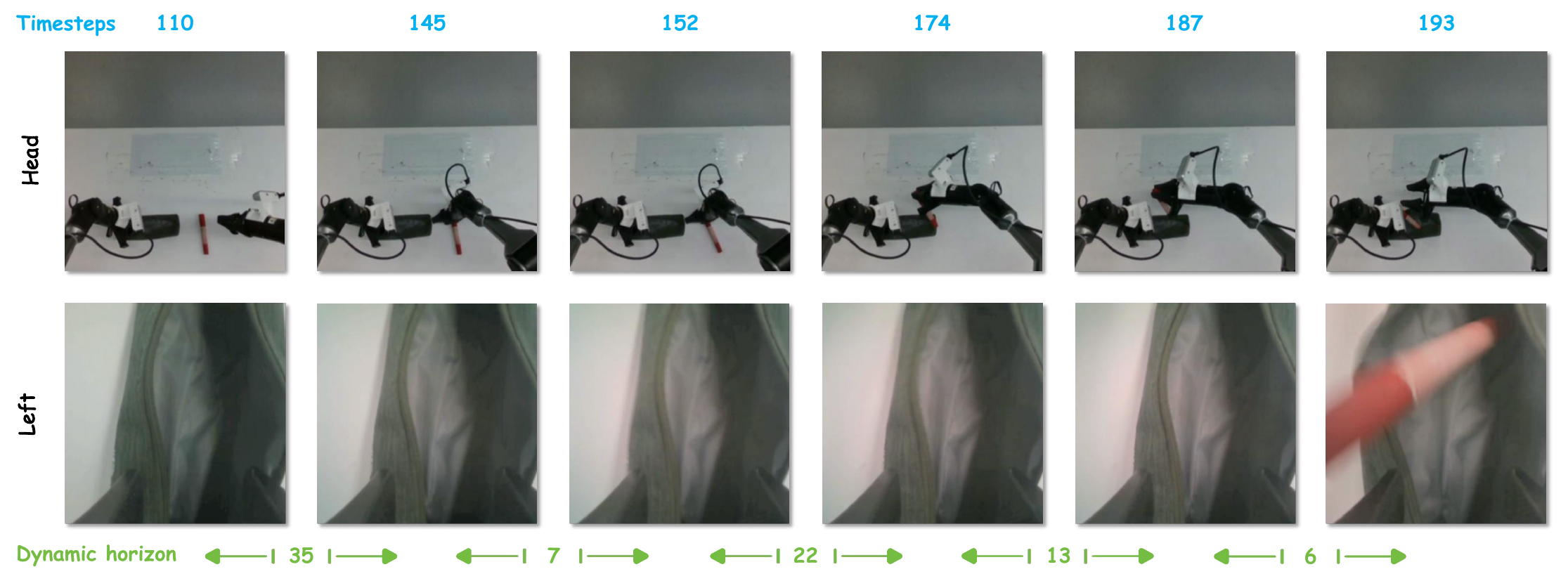}
\caption{
\textbf{Failure case.}
The rollout is from \texttt{put\_pen\_into\_pencil\_case} (top: head
view; bottom: left-wrist view). Blue: policy-query timestep. Green:
horizon length. The pencil case remains only partially opened while the
right arm moves the pen toward it.
}

\label{fig:real_world_failure_rollout}
\end{figure}

Figure~\ref{fig:real_world_failure_rollout} shows a failed
\texttt{put\_pen\_into\_pencil\_case} rollout. The task is strictly
sequential: the left gripper must fully open the pencil case before
the right arm inserts the pen. In this rollout, PACE
selects short horizons (7 and 6 actions) around the insertion phase,
so the policy refreshes its action chunk multiple times from updated
observations. Yet the pencil case remains only partially opened
throughout the segment, and the task fails.

The failure is therefore not due to feedback timing, but to the learned action
distribution of the base policy. PACE already shortens the
execution horizon near insertion and refreshes the chunk from updated
observations, but the refreshed chunks are still produced by the same policy
and do not yield task-completing behavior in this rollout. This case highlights
the boundary of test-time execution control: it can decide when to query the
policy, but it cannot compensate when the policy itself fails to generate
actions sufficient for task completion.

\section{Limitations}
\label{sec:limitations}

PACE cannot resolve failures when the base policy continues to predict incomplete or incorrect actions (Sec.~\ref{sec:real_world_failure_case}).
Its replanning rule can miss phase transitions without clear low-speed valleys.
If no valley is accepted, PACE falls back to the maximum execution horizon.
Effectiveness with policy families and action representations beyond those evaluated here remains to be assessed.
Threshold calibration requires training demonstrations, either per task or pooled across tasks for a shared threshold.
Sharing avoids separate per-task calibration but retains this data requirement.
Earlier replanning can increase policy queries relative to full-chunk execution, raising inference cost even though the selector is inexpensive.

\section{Conclusion}
\label{sec:conclusion}

We showed that the execution horizon is a critical deployment variable with strongly task-dependent and non-monotonic effects on success.
PACE selects execution horizons online from the phase-dependent kinematics of predicted action chunks, using prominent low-speed valleys as candidate replanning boundaries without additional training.
On RoboTwin2.0, PACE improves the average success rate of \(\pi_{0.5}\) over both fixed-horizon and adaptive execution baselines.
Applying PACE to GR00T N1.7 on LIBERO also yields the highest overall success rate among the evaluated strategies, supporting its applicability across different base policies, benchmarks, and action representations.
Real-robot evaluations further demonstrate improvements in task score and success rate over full-chunk execution.
Ablations support motion-profile smoothing and longer training prediction horizons, while a shared threshold retains most gains.
Gains over matched fixed-horizon baselines further support adaptation within a rollout.
PACE offers a practical way to improve task performance through adaptive execution without modifying or retraining the policy.

\clearpage
\begin{ack}
This work was carried out at Peking University and JD Explore Academy.
This work was supported by the Beijing Natural Science Foundation International Scientists Project (Grant No. IS26042).
The authors declare no other competing interests.
\end{ack}

\bibliographystyle{unsrtnat}
\bibliography{ref}

\clearpage
\appendix

\section{PACE implementation details}
\label{app:adaptive_horizon_details}

This appendix specifies the default configuration, online selection algorithm, and demonstration-based threshold calibration for PACE.

\subsection{Default configuration}

Table~\ref{tab:adaptive_horizon_default_config} summarizes the default PACE configuration for the \(\pi_{0.5}\) experiments on RoboTwin2.0.
The GR00T N1.7 configuration on LIBERO is described in Appendix~\ref{app:libero}.

\begin{table}[h]
\centering
\small
\setlength{\tabcolsep}{6pt}
\renewcommand{\arraystretch}{1.1}
\caption{\textbf{Default PACE configuration for \(\pi_{0.5}\) on RoboTwin2.0.}}
\label{tab:adaptive_horizon_default_config}
\begin{tabular}{lc}
\toprule
Component & Default value \\
\midrule
Prediction horizon \(L\) & 50 \\
Maximum execution horizon \(H_{\max}\) & 50 \\
Control frequency \(f_s\) & 50 Hz \\
Speed signal & Joint-space speed \\
Smoothing & Savitzky--Golay filter \\
Filter window / polynomial order & 11 / 3 \\
Minimum valley spacing \(d_{\min}\) & 10 steps \\
Valley score & Absolute peak-height contrast \\
Threshold percentile & 5th percentile from training demonstrations \\
Multi-arm fusion & Earliest accepted boundary across arms \\
\bottomrule
\end{tabular}
\end{table}

The default speed signal is computed from arm joint commands and excludes gripper commands. This matches the policy action space and avoids forward kinematics during deployment.

\subsection{Online horizon selection}

Algorithm~\ref{alg:adaptive_horizon} gives the test-time procedure at one policy query. Let \(\mathcal{B}\) denote the set of controllable arm groups. In our bimanual RoboTwin2.0 experiments, \(\mathcal{B}=\{\mathrm{left},\mathrm{right}\}\). For each arm group, PACE scans low-speed valleys in temporal order and accepts the first valley whose peak-height contrast exceeds the task threshold. The final execution horizon is the earliest accepted boundary across arms. If no valid valley is found, the method falls back to full-chunk execution.

\begin{algorithm}[h]
\caption{PACE at a policy query}
\label{alg:adaptive_horizon}
\begin{algorithmic}[1]
\Require Predicted action chunk \(\mathbf{A}_i\), arm groups \(\mathcal{B}\), maximum horizon \(H_{\max}\), threshold \(\delta_{\mathcal{T}}\)
\Ensure Execution horizon \(h_i\)
\For{each arm group \(b \in \mathcal{B}\)}
    \State Compute speed profile \(\mathbf{v}_i^b\) from \(\mathbf{A}_i\)
    \State Smooth the profile: \(\tilde{\mathbf{v}}_i^b \leftarrow \mathcal{S}(\mathbf{v}_i^b)\)
    \State Detect candidate valleys \(\mathcal{V}_i^b\) with spacing \(d_{\min}\)
    \State Set \(h_i^b \leftarrow H_{\max}\)
    \For{valley \(r \in \mathcal{V}_i^b\) in temporal order}
        \If{\(r > H_{\max}\)}
            \State \textbf{break}
        \EndIf
        \State Compute peak-height contrast \(\Delta_i^b(r)\)
        \If{\(\Delta_i^b(r) \ge \delta_{\mathcal{T}}\)}
            \State \(h_i^b \leftarrow r\)
            \State \textbf{break}
        \EndIf
    \EndFor
\EndFor
\State \Return \(h_i=\min\left(\{h_i^b:b\in\mathcal{B}\}\cup\{H_{\max},L\}\right)\)
\end{algorithmic}
\end{algorithm}

For a single-arm policy or a single-arm task, we simply set \(\mathcal{B}=\{b\}\), where \(b\) is the executed arm. No dummy second arm is introduced. The final horizon then reduces to
\begin{equation}
h_i=\min\{h_i^b,H_{\max},L\}.
\end{equation}
Thus, the single-arm implementation is the same algorithm without the cross-arm minimum. If the single arm has no accepted valley, the fallback remains \(h_i=H_{\max}\).

\subsection{Offline threshold calibration from demonstrations}
\label{app:threshold_calibration}

The only task-specific quantity in PACE is the valley-acceptance
threshold \(\delta_{\mathcal{T}}\). We calibrate this threshold once per task
from training demonstrations and then keep it fixed for evaluation. This
calibration does not use evaluation rollouts, success labels, or fixed-horizon
sweeps.

For each task, we compute speed profiles from 50 training demonstrations using
the same signal type, smoothing filter, and minimum-spacing parameter as the
online controller. Candidate valleys divide each demonstration into motion
segments. For a segment \(s=[u,v]\), we compute its peak-height contrast as
\begin{equation}
\Delta(s)
=
\min\left\{
\max_{u\le j\le v}\tilde v_j-\tilde v_u,
\;
\max_{u\le j\le v}\tilde v_j-\tilde v_v
\right\}.
\end{equation}
We collect positive contrasts across training segments and set the task
threshold to
\begin{equation}
\delta_{\mathcal{T}}
=
\mathrm{Percentile}_{5}
\left(
\{\Delta(s): s \in \text{training segments of task } \mathcal{T}\}
\right).
\end{equation}
The same calibration procedure is used for single-arm and multi-arm settings;
the only difference is the arm set \(\mathcal{B}\) used to construct speed
profiles.

We also pool training demonstrations from all 50 RoboTwin2.0 tasks and apply the same calibration procedure to obtain a shared threshold (Appendix~\ref{app:shared_threshold}).

\section{Simulation protocols and results}
\label{app:sim_full_results}

\subsection{RoboTwin2.0}
\label{app:robotwin_results}

\subsubsection{Evaluation protocol}
\label{app:sim_eval_protocol}

We evaluate all methods on the 50-task RoboTwin2.0 benchmark under the \texttt{demo\_clean} setting. Each task uses one trained \(\pi_{0.5}\) checkpoint with prediction horizon \(L=50\). Within each task, all compared execution schemes share the same checkpoint, 
and only the test-time execution scheme is changed.

For each task, we fine-tune the pretrained \(\pi_{0.5}\) model on 50 \texttt{demo\_clean} demonstrations.
All model parameters are updated using AdamW with a global batch size of 32 and a fixed training seed of 42.
The learning rate is warmed up to \(2.5\times 10^{-5}\) over 1,000 steps and then decayed to \(2.5\times 10^{-6}\) following a cosine schedule.
We evaluate the checkpoint saved at step 30,000, using exponential moving average weights with a decay of 0.99.

We compare PACE against fixed-horizon execution with \(H\in\{5,25,50\}\), AAC, and AutoHorizon.
The \(H=50\) baseline corresponds to full-chunk execution.
For fixed-horizon execution, the robot executes the first \(H\) actions of each predicted chunk before re-querying the policy.
For PACE, the executed prefix is selected by the rule in Appendix~\ref{app:adaptive_horizon_details}.
The AAC and AutoHorizon implementations are described in Appendix~\ref{app:adaptive_baselines}.
In all methods, the executed prefix is clamped by the returned chunk length.

For each task and method, we evaluate seeds \(0,1,2\). For each seed, we run three batches of 100 episodes. Thus, each task-method pair is evaluated over \(3\times 3\times 100=900\) episodes. Success is measured using the RoboTwin2.0 task success predicate. The task-level success rate is averaged over the nine 100-episode runs, and the 50-task average is the equal-weight average over all task-level success rates.

\subsubsection{Full results and aggregate statistics}
\label{app:adaptive_baseline_statistics}

Table~\ref{tab:full_main_sim_results} reports success rates for PACE, fixed-horizon execution, AAC, and AutoHorizon on all 50 RoboTwin2.0 tasks.
The best single global fixed-horizon baseline is \(H=25\), with a 50-task average success rate of \(57.8\%\).
PACE improves the average success rate to \(64.2\%\).

\begingroup
\small
\setlength{\tabcolsep}{3.8pt}
\renewcommand{\arraystretch}{1.05}
\captionsetup{
  width=\linewidth,
  justification=raggedright,
  singlelinecheck=false
}

\begin{xltabular}{\linewidth}{@{}
  >{\raggedright\arraybackslash}X
  cccccc
@{}}

\caption{
\textbf{Results on all 50 RoboTwin2.0 tasks.}
Success rates (\%). Bold marks the highest mean in each row. The final rows report the mean and sample SD of the three seed-level overall success rates.
}
\label{tab:full_main_sim_results} \\
\toprule
Task & \(H=5\) & \(H=25\) & \(H=50\) & AutoHorizon & AAC & PACE \\
\midrule
\endfirsthead

\toprule
Task & \(H=5\) & \(H=25\) & \(H=50\) & AutoHorizon & AAC & PACE \\
\midrule
\endhead

\midrule
\multicolumn{7}{r}{Continued on next page} \\
\endfoot

\bottomrule
\endlastfoot

\texttt{adjust\_bottle} & 99.0 & 98.8 & 96.1 & 98.7 & \textbf{99.7} & 99.6 \\
\texttt{beat\_block\_hammer} & \textbf{81.1} & 67.7 & 66.6 & 71.3 & 64.8 & 80.7 \\
\texttt{blocks\_ranking\_rgb} & 7.3 & 50.0 & 36.1 & \textbf{66.1} & 52.4 & 61.4 \\
\texttt{blocks\_ranking\_size} & 23.9 & 18.6 & 9.2 & 25.9 & 20.2 & \textbf{32.9} \\
\texttt{click\_alarmclock} & \textbf{92.4} & 65.3 & 65.9 & 71.4 & 86.7 & 87.6 \\
\texttt{click\_bell} & \textbf{97.8} & 49.2 & 41.4 & 70.8 & 64.3 & 91.4 \\
\texttt{dump\_bin\_bigbin} & 58.7 & 85.1 & 84.3 & 88.1 & 89.1 & \textbf{93.0} \\
\texttt{grab\_roller} & 89.9 & 96.2 & 96.2 & 96.9 & 97.7 & \textbf{98.6} \\
\texttt{handover\_block} & 3.6 & 33.2 & 52.6 & 43.1 & 43.2 & \textbf{53.1} \\
\texttt{handover\_mic} & 77.2 & 97.7 & 96.8 & 97.7 & 96.8 & \textbf{99.4} \\
\texttt{hanging\_mug} & 16.2 & 15.8 & 16.7 & 20.1 & \textbf{23.6} & 21.8 \\
\texttt{lift\_pot} & 50.9 & 61.2 & 62.0 & 62.2 & 63.1 & \textbf{65.4} \\
\texttt{move\_can\_pot} & 55.4 & \textbf{74.3} & 58.2 & 56.8 & 60.2 & 72.3 \\
\texttt{move\_pillbottle\_pad} & 34.8 & \textbf{49.9} & 29.1 & 44.3 & 46.6 & 45.0 \\
\texttt{move\_playingcard\_away} & 79.2 & \textbf{79.7} & 69.6 & 72.1 & 70.4 & 79.3 \\
\texttt{move\_stapler\_pad} & 12.6 & 11.4 & 7.1 & 13.6 & 10.6 & \textbf{18.6} \\
\texttt{open\_laptop} & 84.9 & 84.8 & 88.8 & \textbf{89.6} & 89.3 & 88.0 \\
\texttt{open\_microwave} & 33.3 & 26.0 & \textbf{60.3} & 55.4 & 52.4 & 45.0 \\
\texttt{pick\_diverse\_bottles} & 13.4 & 23.9 & 24.1 & 26.4 & 27.9 & \textbf{35.7} \\
\texttt{pick\_dual\_bottles} & 13.3 & 45.3 & 38.7 & 41.6 & \textbf{48.9} & 48.2 \\
\texttt{place\_a2b\_left} & 54.6 & 52.7 & 43.2 & 46.4 & 50.4 & \textbf{55.9} \\
\texttt{place\_a2b\_right} & 20.0 & 47.8 & 36.8 & 43.1 & 46.7 & \textbf{54.6} \\
\texttt{place\_bread\_basket} & \textbf{55.3} & 53.9 & 30.3 & 42.6 & 46.2 & 49.1 \\
\texttt{place\_bread\_skillet} & 38.4 & 39.8 & 29.0 & 33.2 & \textbf{42.0} & 37.0 \\
\texttt{place\_burger\_fries} & 88.9 & 89.3 & 85.8 & 87.1 & \textbf{91.9} & 91.3 \\
\texttt{place\_can\_basket} & 18.1 & 56.8 & 38.2 & 52.1 & \textbf{66.3} & 53.6 \\
\texttt{place\_cans\_plasticbox} & 85.1 & 83.0 & 81.8 & 85.2 & 85.8 & \textbf{90.6} \\
\texttt{place\_container\_plate} & 88.3 & 81.7 & 91.6 & 91.2 & 89.1 & \textbf{96.1} \\
\texttt{place\_dual\_shoes} & 19.9 & \textbf{40.6} & 33.7 & 31.0 & 40.3 & 37.3 \\
\texttt{place\_empty\_cup} & \textbf{59.4} & 48.1 & 45.8 & 51.8 & 45.9 & 58.8 \\
\texttt{place\_fan} & 35.3 & 34.0 & 25.8 & 31.4 & 35.7 & \textbf{39.1} \\
\texttt{place\_mouse\_pad} & 4.3 & 14.1 & 12.8 & 16.3 & 15.1 & \textbf{17.8} \\
\texttt{place\_object\_basket} & 35.9 & 59.4 & 58.0 & 67.8 & 64.8 & \textbf{70.2} \\
\texttt{place\_object\_scale} & 23.8 & 29.4 & 18.9 & 23.8 & 24.8 & \textbf{40.7} \\
\texttt{place\_object\_stand} & \textbf{64.2} & 57.3 & 53.4 & 56.6 & 54.9 & 59.2 \\
\texttt{place\_phone\_stand} & 52.9 & \textbf{59.9} & 54.7 & 54.7 & 55.4 & 58.7 \\
\texttt{place\_shoe} & 43.9 & \textbf{64.4} & 56.1 & 58.6 & 63.0 & 62.8 \\
\texttt{press\_stapler} & \textbf{68.2} & 60.6 & 56.0 & 59.8 & 61.6 & 61.8 \\
\texttt{put\_bottles\_dustbin} & 0.3 & 79.3 & 70.8 & 74.3 & 76.6 & \textbf{88.9} \\
\texttt{put\_object\_cabinet} & 7.6 & 51.9 & 50.6 & 50.6 & 48.7 & \textbf{57.4} \\
\texttt{rotate\_qrcode} & 48.9 & 69.9 & 66.1 & 70.2 & 68.4 & \textbf{71.7} \\
\texttt{scan\_object} & 20.9 & \textbf{43.1} & 28.9 & 39.7 & 39.9 & 40.3 \\
\texttt{shake\_bottle} & 95.2 & 95.7 & 96.7 & 96.7 & 96.2 & \textbf{98.4} \\
\texttt{shake\_bottle\_horizontally} & 97.3 & 96.8 & 96.9 & 97.1 & 98.6 & \textbf{98.8} \\
\texttt{stack\_blocks\_three} & 39.7 & 43.2 & 28.7 & 54.6 & 39.1 & \textbf{55.2} \\
\texttt{stack\_blocks\_two} & 12.4 & 80.1 & 63.9 & \textbf{82.9} & 77.3 & 82.3 \\
\texttt{stack\_bowls\_three} & 45.0 & 61.0 & 67.6 & 70.8 & 64.0 & \textbf{79.8} \\
\texttt{stack\_bowls\_two} & 88.7 & 88.4 & 85.1 & 91.0 & 88.2 & \textbf{95.6} \\
\texttt{stamp\_seal} & \textbf{67.4} & 34.2 & 24.3 & 37.0 & 32.1 & 47.8 \\
\texttt{turn\_switch} & 34.4 & 39.4 & 37.1 & 41.8 & 36.7 & \textbf{42.6} \\
\midrule
\textbf{Mean} & 48.79 & 57.80 & 53.36 & 59.03 & 59.07 & \textbf{64.20} \\
\textbf{Seed SD} & 0.89 & 1.49 & 0.44 & 0.53 & 0.56 & 0.88 \\
\end{xltabular}
\endgroup

For each evaluation seed, we average the three batches within each task and then average equally over all 50 tasks.
The final two rows of Table~\ref{tab:full_main_sim_results} summarize the resulting three overall success rates.
Each task uses the same checkpoint across seeds, so the SD reflects evaluation variability.

PACE achieves a higher mean success rate than AutoHorizon on 46 of the 50 tasks and than AAC on 39 of the 50 tasks.

Using unrounded success rates, PACE improves overall success by \(6.40\) percentage points over fixed \(H=25\) and by \(5.13\) points over AAC.
Paired bootstraps over the 50 task-level mean differences yield 95\% confidence intervals of \([+4.21,+8.86]\) and \([+3.22,+7.10]\), respectively.
Against AutoHorizon, PACE gains \(5.18\) percentage points, with a 95\% confidence interval of \([+3.62,+6.78]\) from a paired hierarchical bootstrap.
These confidence intervals describe uncertainty in the gains over each baseline, while the SD in Table~\ref{tab:full_main_sim_results} describes variability across evaluation seeds.

\subsubsection{Adaptive baseline implementations}
\label{app:adaptive_baselines}

\paragraph{Shared experimental setting.}
AutoHorizon and AAC use the same task-specific \(\pi_{0.5}\) checkpoints as the fixed-horizon baselines and PACE, without additional policy training.
Both use ten flow-sampling steps to predict 50-step chunks of decoded 14-D actions, with six absolute joint-position commands and one gripper command per arm.
They follow the evaluation protocol in Appendix~\ref{app:sim_eval_protocol}.
For each baseline, the three batches within a task and evaluation seed reuse the same set of 100 evaluation scenes.
Their selector settings are fixed across tasks and are not tuned using evaluation success labels.

\paragraph{AutoHorizon.}
We implement AutoHorizon~\citep{wang2026vla} in JAX following the official implementation.
The selector uses model-internal action-token self-attention from denoising step 3 and applies the bidirectional soft-pointer rule with \(\tau=0.3\), entropy quantile \(0.9\), and run length \(1\).
The robot executes the selected prefix of the decoded action chunk, discards the remaining actions, and queries the policy again using a new observation.

\paragraph{AAC adapted to bimanual joint-space actions.}
We adapt AAC~\citep{liang2026adaptive} to RoboTwin's bimanual joint-space actions and sample 20 candidate chunks per policy query.
At each prediction step, we estimate Gaussian entropy from the six normalized joint commands of each arm across the candidates, with covariance regularization \(10^{-6}I\).
For each gripper, we binarize its decoded commands at \(0.5\) and compute Bernoulli entropy.

Movement is measured from the decoded absolute joint targets relative to the current state, summing the arm-wise displacements and weighting gripper toggles by \(0.2\).
We use the movement threshold of \(3.0\) reported in the AAC supplementary material.
The final execution horizon is the larger of the horizon selected by AAC's prefix-entropy turning-point rule and the shortest prefix satisfying the movement constraint.

To generate AAC's 20 candidate chunks, our \(\pi_{0.5}\) implementation encodes the image--language--state prefix once per query and reuses its KV cache across candidates.
The action head processes candidate 0 separately and the remaining 19 candidates as a batch.
This grouping preserves candidate 0's agreement with standard \(\pi_{0.5}\) sampling.
The robot executes the selected prefix of candidate 0.

\subsection{GR00T N1.7 on LIBERO}
\label{app:libero}

\paragraph{Policy and evaluation protocol.}
We evaluate PACE using the official NVIDIA/LeRobot GR00T N1.7 checkpoints~\citep{nvidia2026grootn17} on LIBERO-Spatial, LIBERO-Object, LIBERO-Goal, and LIBERO-10~\citep{liu2023libero}.
The four suites contain 10 tasks each, for 40 tasks in total.
Within each suite, all execution strategies use the same checkpoint without retraining or modifying the base policy.
We evaluate each task on 50 initial states for each of three evaluation seeds, yielding 150 episodes per task and 6,000 episodes per strategy.

For each seed, Overall is the equal-weight mean success rate across all 40 tasks.
We report the mean and sample standard deviation across the three seeds for each suite and for Overall.

\paragraph{Adapting PACE.}
GR00T N1.7 predicts 16-step Cartesian-delta chunks, whereas the \(\pi_{0.5}\) policies evaluated on RoboTwin2.0 predict 50-step joint-space chunks.
PACE constructs the motion profile from the norm of the predicted translational action deltas and applies the same valley-based horizon-selection rule.
PACE calibrates its acceptance thresholds from LIBERO training demonstrations using valley contrasts normalized by peak speed.

We compare PACE with fixed horizons \(H=2,4,8,16\) and AutoHorizon under the same evaluation protocol.
The \(H=16\) baseline executes the full predicted chunk.
AutoHorizon follows its published horizon-selection rule adapted to GR00T N1.7.

\begin{table}[ht]
\centering
\small
\setlength{\tabcolsep}{5pt}
\renewcommand{\arraystretch}{1.18}
\caption{\textbf{GR00T N1.7 results on LIBERO.}
Success rates (\%), reported as mean \(\pm\) sample SD across three evaluation seeds.
Bold marks the highest mean in each column.}
\label{tab:libero}
\begin{tabular}{@{}lccccc@{}}
\toprule
\makecell[l]{Execution\\strategy} & Spatial & Object & Goal & LIBERO-10 & Overall \\
\midrule
\(\mathrm{H}=2\) & \(97.20\pm0.92\) & \(91.47\pm0.50\) & \(85.87\pm1.17\) & \(77.73\pm1.33\) & \(88.07\pm0.28\) \\
\(\mathrm{H}=4\) & \(\mathbf{97.67}\pm0.31\) & \(94.33\pm0.90\) & \(93.13\pm1.62\) & \(83.33\pm1.92\) & \(92.12\pm0.30\) \\
\(\mathrm{H}=8\) & \(96.87\pm0.90\) & \(96.20\pm0.72\) & \(96.00\pm0.53\) & \(88.00\pm0.87\) & \(94.27\pm0.37\) \\
\(\mathrm{H}=16\) & \(96.27\pm0.99\) & \(97.00\pm0.92\) & \(96.53\pm1.01\) & \(88.07\pm0.31\) & \(94.47\pm0.23\) \\
\midrule
AutoHorizon & \(96.13\pm0.42\) & \(96.67\pm0.31\) & \(95.80\pm0.80\) & \(\mathbf{91.07}\pm1.45\) & \(94.92\pm0.44\) \\
\rowcolor{blue!6}\textbf{PACE} & \(97.00\pm0.72\) & \(\mathbf{97.73}\pm0.23\) & \(\mathbf{96.93}\pm0.90\) & \(89.67\pm1.17\) & \(\mathbf{95.33}\pm0.18\) \\
\bottomrule
\end{tabular}
\end{table}

\paragraph{Results and interpretation.}
PACE achieves the highest overall success rate among the evaluated strategies (Table~\ref{tab:libero}).
Using unrounded rates, it exceeds \(H=16\), the strongest global fixed-horizon baseline, by 0.87 percentage points.
A paired hierarchical bootstrap stratified by suite gives a 95\% confidence interval of \([+0.10,+1.70]\) percentage points for this gain.
Its overall success rate is also 0.42 percentage points higher than AutoHorizon.
Across individual suites, PACE leads on Object and Goal, fixed \(H=4\) on Spatial, and AutoHorizon on LIBERO-10.

\paragraph{Policy-query and inference overhead.}
We measure inference latency on a dedicated NVIDIA H20 using the LIBERO setup described above.
Across the four suite-specific GR00T N1.7 checkpoints, model-forward latency ranges from \(127.55\) to \(130.06\) ms per query.
PACE's horizon selector takes \(0.0566\pm0.0046\) ms per query, approximately \(0.044\%\) of the time required for a policy forward pass.
The total selector time per successful episode is \(0.595\pm0.001\) ms.

Table~\ref{tab:libero_overhead} reports policy queries and total inference time per successful episode for each execution strategy.
These metrics are computed separately for each evaluation seed using that strategy's successful episodes.
We report the mean and sample standard deviation across the three seeds.
The reported times cover inference computation only.

\begin{table}[ht]
\centering
\small
\setlength{\tabcolsep}{10pt}
\renewcommand{\arraystretch}{1.18}
\caption{\textbf{Policy queries and inference time with GR00T N1.7 on LIBERO.}
Total inference time includes policy forward passes and, for PACE, horizon selection.}
\label{tab:libero_overhead}
\begin{tabular}{@{}lcc@{}}
\toprule
\makecell[l]{Execution\\strategy} &
\makecell{Queries per\\successful episode} &
\makecell{Total inference time per\\successful episode (s)} \\
\midrule
\(\mathrm{H}=2\) & \(78.21\pm0.43\) & \(10.075\pm0.056\) \\
\(\mathrm{H}=4\) & \(38.49\pm0.10\) & \(4.959\pm0.013\) \\
\(\mathrm{H}=8\) & \(19.05\pm0.08\) & \(2.455\pm0.010\) \\
\(\mathrm{H}=16\) & \(9.69\pm0.01\) & \(1.248\pm0.002\) \\
\midrule
\rowcolor{blue!6}\textbf{PACE} & \(10.53\pm0.02\) & \(1.357\pm0.003\) \\
\bottomrule
\end{tabular}
\end{table}

Compared with full-chunk execution (\(H=16\)), PACE improves success by \(0.87\) percentage points while increasing total inference time per successful episode by \(8.66\%\).
This increase comes almost entirely from the \(0.83\) additional policy queries per successful episode, with horizon selection contributing little.
All differences are computed from unrounded measurements.

\section{Ablation and robustness studies}
\label{app:ablation_details}

The phase-profile and valley-selection studies follow the full RoboTwin2.0 protocol in Appendix~\ref{app:sim_eval_protocol}: 50 tasks and 900 episodes per task across three seeds.
Average horizon denotes the mean planned execution length.
The training-horizon study requires retraining and uses six representative tasks, seed \(0\), and three 100-episode batches per setting.

\subsection{Phase-profile construction}
\label{app:phase_profile_ablation}

We compare Cartesian and joint-space speed, with and without Savitzky--Golay smoothing, while keeping the valley-selection rule and calibration percentile fixed.
The Cartesian variant computes end-effector linear speed using forward kinematics; the raw variants detect valleys without smoothing.
Table~\ref{tab:design_analysis} (left) reports the results.
Smoothing improves both representations, consistent with suppressing spurious short-range valleys.
The smoothed Cartesian variant has a slight success-rate advantage, while the default smoothed joint-space variant produces longer execution horizons and avoids forward kinematics at test time.

\subsection{Training prediction horizon}
\label{app:training_prediction_horizon}

We study how the policy prediction horizon used during training affects
fixed-horizon execution at test time. Here \(H_{\mathrm{train}}\) denotes
the supervised action-chunk length during policy training, while
\(H_{\mathrm{eval}}\) denotes the fixed executed prefix length at
evaluation. Entries with \(H_{\mathrm{train}} < H_{\mathrm{eval}}\) are
infeasible because the policy does not predict enough actions to execute
the requested prefix. We sweep both axes on a finer grid, from \(5\) to
\(50\) in steps of \(5\), using six representative tasks, seed \(0\), and
three 100-episode batches per setting.

\begin{figure}[h]
\centering
\includegraphics[width=0.85\linewidth]{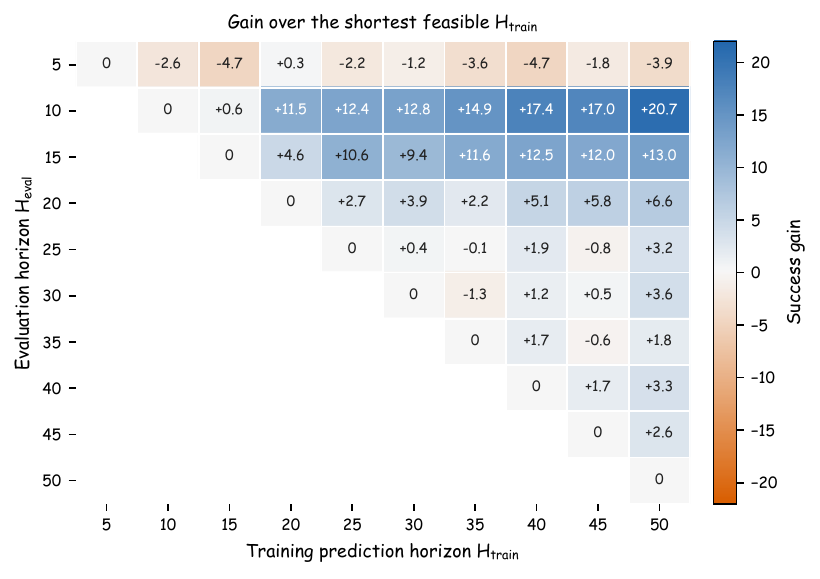}
\caption{
\textbf{Training prediction horizon ablation.}
Each cell shows the success gain of training horizon
\(H_{\mathrm{train}}\) (columns) relative to the shortest feasible
training horizon for a given evaluation horizon \(H_{\mathrm{eval}}\)
(rows), i.e., \(H_{\mathrm{train}}{=}H_{\mathrm{eval}}\). Blue indicates
a gain, orange a loss; the diagonal is zero by construction. Cells
below the diagonal are infeasible and left blank.
}
\label{fig:app_training_horizon_heatmap}
\end{figure}

Figure~\ref{fig:app_training_horizon_heatmap} shows that, for \(H_{\mathrm{eval}}\geq10\), increasing \(H_{\mathrm{train}}\) generally improves success, with gains reaching \(+20.7\), \(+13.0\), and \(+6.6\) points at \(H_{\mathrm{eval}}=10,15,20\), respectively.
Very short execution is an exception: at \(H_{\mathrm{eval}}=5\), most gains lie between \(-4.7\) and \(0\).
For \(H_{\mathrm{eval}}>20\), gains remain below \(+7\) points and become more variable, including near-zero and slightly negative values.
Longer training horizons thus benefit short but non-trivial execution prefixes most, motivating the default \(H_{\mathrm{train}}=50\).

\subsection{Valley-selection sensitivity}
\label{app:valley_sensitivity}

We vary the minimum spacing between candidate valleys and the calibration percentile
\(\rho\) used to set the contrast threshold \(\delta\), keeping all other settings fixed. Larger
\(\rho\) produces a stricter threshold, which accepts fewer valleys and
therefore tends to increase the selected execution horizon.

\begin{table}[h]
\centering
\small
\setlength{\tabcolsep}{6pt}
\renewcommand{\arraystretch}{1.12}
\caption{
\textbf{Sensitivity to valley-selection parameters.}
All results use the smoothed joint-space phase profile and are averaged
over all 50 RoboTwin2.0 tasks.
}
\label{tab:app_valley_sensitivity}
\begin{tabular}{cccc}
\toprule
Min. spacing & Percentile \(\rho\) & Succ. & Avg. Horizon \\
\midrule
5  & 0  & \best{64.3} & 18.1 \\
5  & 5  & 63.5 & 23.4 \\
5  & 10 & 63.3 & 24.6 \\
5  & 25 & 62.5 & 31.9 \\
\midrule
10 & 0  & 64.0 & 19.8 \\
\oursrow 10 & 5$^\dagger$ & 64.2 & 24.3 \\
10 & 10 & 63.0 & 25.5 \\
10 & 25 & 62.0 & 32.9 \\
\bottomrule
\end{tabular}
\end{table}

PACE is not sensitive to a single narrowly tuned setting.
Several nearby configurations achieve similar success rates. Very strict
thresholds, such as \(\rho=25\), select substantially longer execution
windows and reduce success, suggesting that rejecting too many phase
valleys delays feedback in contact-sensitive stages. We use minimum
spacing \(10\) and \(\rho=5\) as the default because it provides high
success while keeping the average execution horizon moderate.

\subsection{Shared versus per-task threshold calibration}
\label{app:shared_threshold}

To evaluate whether PACE can use one valley-acceptance threshold across tasks, we compare per-task and shared-threshold calibration.
The default setting calibrates a separate threshold from each task's training demonstrations.
For the shared setting, we pool training demonstrations from all 50 RoboTwin2.0 tasks and apply the same calibration procedure (Appendix~\ref{app:threshold_calibration}), retaining the speed-profile construction and 5th-percentile rule.
The resulting threshold is used unchanged across all 50 tasks.
All compared execution strategies use the same task-specific \(\pi_{0.5}\) checkpoints, and all other PACE settings and evaluation conditions remain unchanged.

Following the evaluation protocol in Appendix~\ref{app:sim_eval_protocol}, we compute the equal-weight mean success rate over all 50 tasks for each of three evaluation seeds.
Table~\ref{tab:shared_threshold} reports the mean and sample standard deviation of these three averages.

Shared-threshold PACE improves success over fixed \(H=25\) by \(4.64\) percentage points, with a 95\% confidence interval of \([+2.61,+6.99]\).
Per-task calibration provides an additional gain of approximately \(1.8\) points over the shared threshold, with a 95\% confidence interval of \([+1.18,+2.35]\).
We compute each interval with a paired bootstrap over the 50 task-level differences using unrounded mean success rates.
These results show that a single shared threshold preserves most of PACE's improvement over fixed \(H=25\).

\subsection{Calibration stability and trajectory sensitivity}
\label{app:calibration_stability}

\paragraph{Data and sampling.}
To assess the stability of PACE's horizon decisions, we conduct an offline analysis of trajectories from DROID~\citep{khazatsky2024droid}, AgiBot World Beta~\citep{bu2025agibot}, RoboMIND~\citep{wu2025robomind}, and RoboCOIN~\citep{robocoin2025}.
We use stratified sampling by task and robot configuration, covering 1,384 eligible strata across 31 dataset-defined robot configurations.
For each stratum, we use 50 episodes for calibration and up to 50 held-out episodes for evaluation.
The resulting sample contains 135,046 distinct episodes and 263,319 fixed 50-step evaluation windows.

\paragraph{Calibration resampling.}
For each stratum, we first compute a reference valley-acceptance threshold from its 50 calibration demonstrations using the fifth-percentile rule.
We then resample these demonstrations with replacement 500 times and recompute the threshold for each sample.
On the same held-out evaluation windows, we compare the horizons selected by each recalibrated threshold with those selected by the reference threshold.

\paragraph{Trajectory perturbation.}
To assess sensitivity to trajectory fluctuations, we double the high-frequency residual in the joint-position action sequences.
This residual is the difference between each sequence and its trend obtained with a five-step, second-order Savitzky--Golay filter.
We perturb both the calibration trajectories and the held-out evaluation windows, and recalibrate the threshold on the perturbed demonstrations.
On the perturbed evaluation windows, we measure horizon agreement within five control steps of the original decisions.

\begin{table}[ht]
\centering
\small
\setlength{\tabcolsep}{8pt}
\renewcommand{\arraystretch}{1.18}
\caption{\textbf{Offline stability of PACE's horizon decisions.}
For each metric, we average equally over task/configuration strata within each dataset family and then equally across the four families.}
\label{tab:calibration_stability}
\begin{tabular}{@{}llr@{}}
\toprule
Test & Metric & Result \\
\midrule
\makecell[l]{Calibration\\resampling} &
\makecell[l]{Threshold coefficient of variation\\(dataset-family-balanced)} & \(18.7\%\) \\
& Identical horizon & \(99.08\%\) \\
& Absolute horizon shift \(>5\) steps & \(0.78\%\) \\
& Mean absolute horizon shift & \(0.14\) steps \\
\midrule
\makecell[l]{High-frequency\\residual doubled} & Absolute horizon shift \(\leq5\) steps & \(80.6\%\) \\
\bottomrule
\end{tabular}
\end{table}

Table~\ref{tab:calibration_stability} reports horizon consistency under calibration resampling and trajectory perturbation.
On a fixed evaluation window, threshold changes leave the horizon unchanged whenever the same earliest valley remains accepted or all candidates remain rejected.

\section{Execution-horizon analysis}
\label{app:fixed_horizon_sweep}

\subsection{Fixed-horizon sweeps}
\label{app:sweep_protocol}

For each of the six representative tasks in Sec.~\ref{sec:fixed_horizon_sweep}, we evaluate every integer fixed horizon \(H\in\{1,\ldots,50\}\).
Each fixed-horizon setting uses seed \(0\) and three 100-episode batches, giving 300 episodes per horizon.
PACE uses three seeds with three 100-episode batches per seed, giving 900 episodes per task.
Its mean execution horizon is averaged over all selected execution windows.
The best swept horizon \(H^\star\) is selected using test-task outcomes and serves as a diagnostic reference within fixed-horizon schedules, not a deployable selection rule.

\begin{table}[h]
\centering
\small
\setlength{\tabcolsep}{6pt}
\renewcommand{\arraystretch}{1.12}
\caption{
\textbf{Diagnostic fixed-horizon sweep summary.}
\(H^\star\) denotes the best fixed horizon in the sweep, with ties listed together.
All success rates are percentages.
}
\label{tab:app_fixed_horizon_sweep_summary}
\begin{tabular}{lccc}
\toprule
Task & \(H^\star\) & Fixed at \(H^\star\) & PACE \\
\midrule
Pick Dual Bottles & 14 & 56.7 & 48.2 \\
Grab Roller & 7, 9, 32 & 97.7 & 98.6 \\
Place A2B Right & 11 & 53.3 & 54.6 \\
Dump Bin Bigbin & 37 & 90.3 & 93.0 \\
Put Bottles Dustbin & 18 & 88.7 & 88.9 \\
Stack Bowls Two & 8 & 95.3 & 95.6 \\
\bottomrule
\end{tabular}
\end{table}

Table~\ref{tab:app_fixed_horizon_sweep_summary} shows substantial task dependence in \(H^\star\), ranging from short horizons such as \(7\) or \(8\) to \(37\).
Grab Roller has three tied best horizons, illustrating the non-monotonic effect of execution length.

\subsection{Matched execution horizons}
\label{app:matched_horizon_details}

We obtain the matched-horizon results in Table~\ref{tab:matched_horizon_main} from the sweep by selecting the fixed horizon nearest to PACE's mean execution horizon on each task.
PACE achieves a higher observed success rate on all six tasks, with gains of \(1.6\)--\(10.6\) percentage points computed from unrounded values.
These results suggest that matching the mean execution horizon alone is insufficient to reproduce PACE's performance.

Pick Dual Bottles illustrates the difference between the two fixed-horizon comparisons.
PACE achieves \(48.2\%\) success, exceeding the matched fixed horizon \(H=26\) (\(46.7\%\)) but remaining below the best swept horizon \(H^\star=14\) (\(56.7\%\)).

In the seed-0 evaluation, PACE achieved \(43.7\%\) success, compared with \(46.0\%\) for fixed \(H=25\), with 300 episodes per method.
Across all three seeds (900 episodes per method), the corresponding success rates are \(48.2\%\) and \(45.3\%\) (Appendix~\ref{app:sim_full_results}), reversing the seed-0 ordering.

\subsection{Phase alignment}
\label{app:phase_alignment}

To examine whether PACE replans near motion-phase transitions, we compare its executed valley-triggered boundaries with independently annotated transitions.
The analysis covers 89 successful rollouts from three RoboTwin2.0 tasks: Place A2B Right, Place Phone Stand, and Pick Dual Bottles.

\paragraph{Blind annotation and matching.}
Annotators identified motion-phase transitions from paired head-camera and external-observer videos without access to PACE's replanning boundaries or boundary overlays.
After the annotations were finalized, we extracted the executed valley-triggered boundaries from the rollout logs.
We matched boundaries one-to-one with annotated transitions within \(\pm5\) control steps (\(\pm0.1\)\,s at 50\,Hz).
We chose the matching with the largest number of pairs, breaking ties by the smallest total temporal distance.
The one-to-one constraint prevents multiple boundaries from being matched to the same transition.

Of the 391 executed replanning boundaries, six were \(H_{\max}\) fallbacks and were counted separately.
The primary alignment metric uses the remaining 385 valley-triggered boundaries.

\paragraph{Metrics and temporal controls.}
Alignment precision is the fraction of executed valley-triggered boundaries matched to annotated transitions.
Transition coverage is the fraction of annotated transitions matched to a boundary.
For each task, we obtain 95\% confidence intervals for PACE precision by bootstrapping the audited successful rollouts at the trajectory level.

To assess whether the observed alignment exceeds chance levels, we compare PACE with two offline controls that retain the same number of boundaries.
The first samples boundary times uniformly at random with a minimum spacing of 10 control steps.
The second circularly shifts PACE's boundaries while preserving their relative spacing.
Each control is evaluated over 10,000 randomizations.

\begin{table}[ht]
\centering
\small
\setlength{\tabcolsep}{3.5pt}
\renewcommand{\arraystretch}{1.18}
\caption{\textbf{Phase alignment on successful rollouts.}
The last three columns report alignment precision (\%). Overall is computed from pooled counts across tasks.}
\label{tab:phase_alignment}
\begin{tabular}{@{}lccccc@{}}
\toprule
Task & \makecell{Annotated\\transitions}
& \makecell{Aligned /\\valleys}
& \makecell{PACE precision\\(95\% CI)}
& \makecell{Uniform\\random}
& \makecell{Circular\\shift} \\
\midrule
Place A2B Right & 182 & 135 / 172 & 78.5 [71.1, 85.1] & 45.9 & 46.1 \\
Place Phone Stand & 145 & 96 / 123 & 78.0 [72.1, 84.0] & 43.7 & 43.8 \\
Pick Dual Bottles & 120 & 76 / 90 & 84.4 [78.5, 90.2] & 41.9 & 42.3 \\
\midrule
Overall & 447 & 307 / 385 & 79.7 & -- & -- \\
\bottomrule
\end{tabular}
\end{table}

\paragraph{Results.}
Table~\ref{tab:phase_alignment} shows that 307 of 385 valley-triggered boundaries align with annotated motion-phase transitions, yielding 79.7\% precision.
These matches cover 68.7\% of the 447 annotated transitions.
Matched boundaries lie an average of 2.3 control steps from their corresponding transitions.
PACE achieves higher alignment precision than both the uniform-random and circular-shift controls on all three tasks, with \(p\leq0.0001\) in all six randomization tests.

\paragraph{Interpretation.}
The observed alignment on successful rollouts from these three tasks supports PACE's use of prominent low-speed valleys as cues for candidate replanning boundaries.
Assessing the effect of individual boundary choices on task success would require controlled interventions on boundary timing.

\section{Real-robot experiment details}
\label{app:real_world_details}

\subsection{Tasks, scenes, and scoring rules}
\label{app:real_world_settings}
\label{app:real_world_scoring}

Within each task, all execution strategies use the same fine-tuned \(\pi_{0.5}\) checkpoint and differ only in the test-time execution rule.
Table~\ref{tab:app_real_world_task_summary} summarizes the robots, demonstrations, and trial counts.

\begin{table}[h]
\centering
\small
\setlength{\tabcolsep}{5.5pt}
\renewcommand{\arraystretch}{1.12}
\caption{
\textbf{Real-robot task and data summary.}
Demos denotes the number of demonstration episodes used for fine-tuning.
}
\label{tab:app_real_world_task_summary}
\begin{tabular}{@{}lccc@{}}
\toprule
Task & Robot & Demos & Trials / method \\
\midrule
\texttt{stack\_bowls} & ALOHA & 1{,}047 & 30 \\
\texttt{put\_pen\_into\_pencil\_case} & ALOHA & 1{,}220 & 30 \\
\texttt{place\_object\_on\_plate} & Franka & 301 & \(5\times20\) \\
\bottomrule
\end{tabular}
\end{table}

\begin{figure}[h]
\centering
\begin{minipage}[t]{0.48\linewidth}
    \centering
    \includegraphics[width=\linewidth]{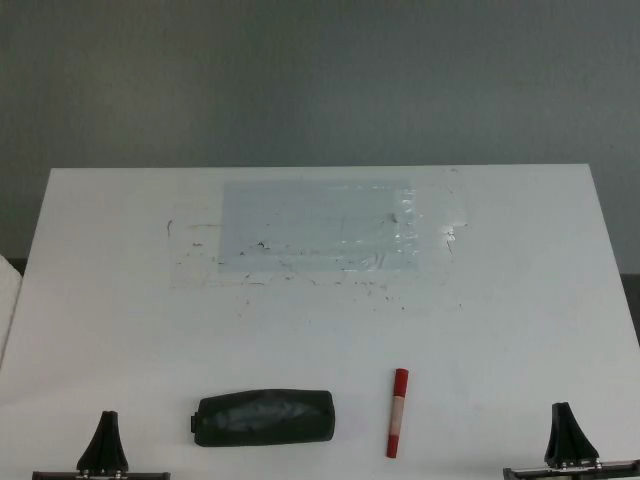}
    \vspace{2pt}
    \small \texttt{put\_pen\_into\_pencil\_case}
\end{minipage}
\hfill
\begin{minipage}[t]{0.48\linewidth}
    \centering
    \includegraphics[width=\linewidth]{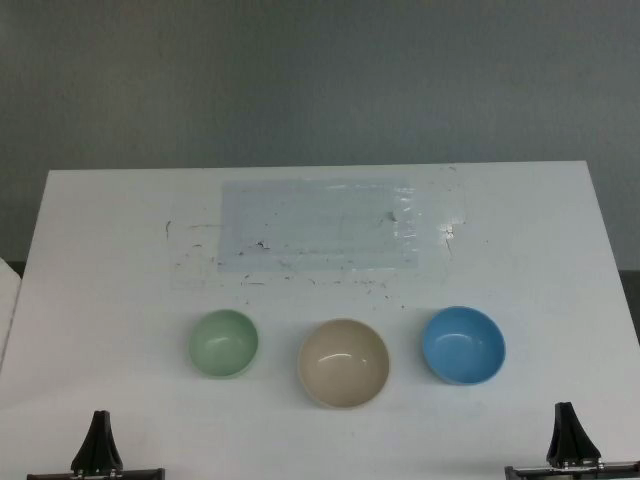}
    \vspace{2pt}
    \small \texttt{stack\_bowls}
\end{minipage}
\caption{
\textbf{Initial frames of the RoboChallenge tasks.}
Left: \texttt{put\_pen\_into\_pencil\_case}. Right:
\texttt{stack\_bowls}. Both tasks use an ALOHA robot.
}
\label{fig:app_robochallenge_scenes}
\end{figure}

\paragraph{RoboChallenge.}
We evaluate \texttt{stack\_bowls} and \texttt{put\_pen\_into\_pencil\_case} on RoboChallenge Table30 V1, following its task prompts, reset procedures, and grading rules.
Figure~\ref{fig:app_robochallenge_scenes} shows the initial scenes.
We report the official partial-credit Score and full-completion rate Succ., using the scoring steps below.

\paragraph{\texttt{stack\_bowls}.}
\begin{itemize}
\item The task is to nest the bowls from largest to smallest.
\item \(2.0 \times 2\) points: pick up a bowl with the gripper.
\item \(2.5 \times 2\) points: successfully stack the bowls.
\item \(1.0\) point: retract the gripper to its original position.
\end{itemize}

\paragraph{\texttt{put\_pen\_into\_pencil\_case}.}
\emph{Note:} The official RoboChallenge description labels the pencil-case
opening step with the opposite arm; in the evaluation setup and rollout videos,
the left gripper opens the pencil case while the right arm manipulates the pen.
We therefore report the scoring steps using the observed arm assignment.
\begin{itemize}
\item The task is to place the pen on the table into the pencil case.
\item \(2.0\) points: left gripper reaches and opens the pencil case.
\item \(2.0\) points: the pencil case is sufficiently open for insertion.
\item \(2.0\) points: right arm picks up the pen.
\item \(3.0\) points: right arm places the pen into the pencil case.
\item \(1.0\) point: reset both arms.
\end{itemize}

\begin{figure}[h]
\centering
\begin{minipage}[t]{0.48\linewidth}
    \centering
    \includegraphics[width=\linewidth]{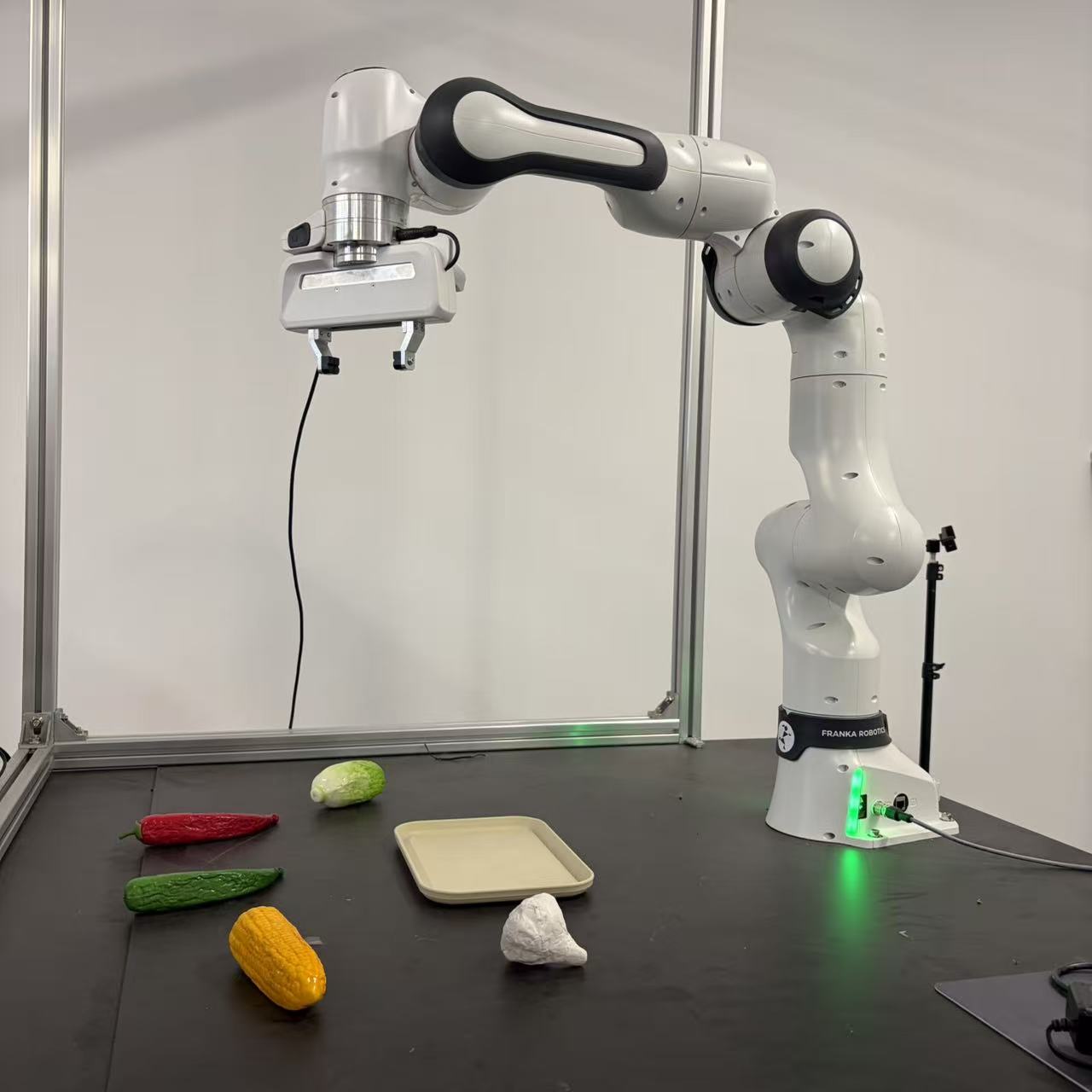}
    \vspace{2pt}
    \small Object and plate layout
\end{minipage}
\hfill
\begin{minipage}[t]{0.48\linewidth}
    \centering
    \includegraphics[width=\linewidth]{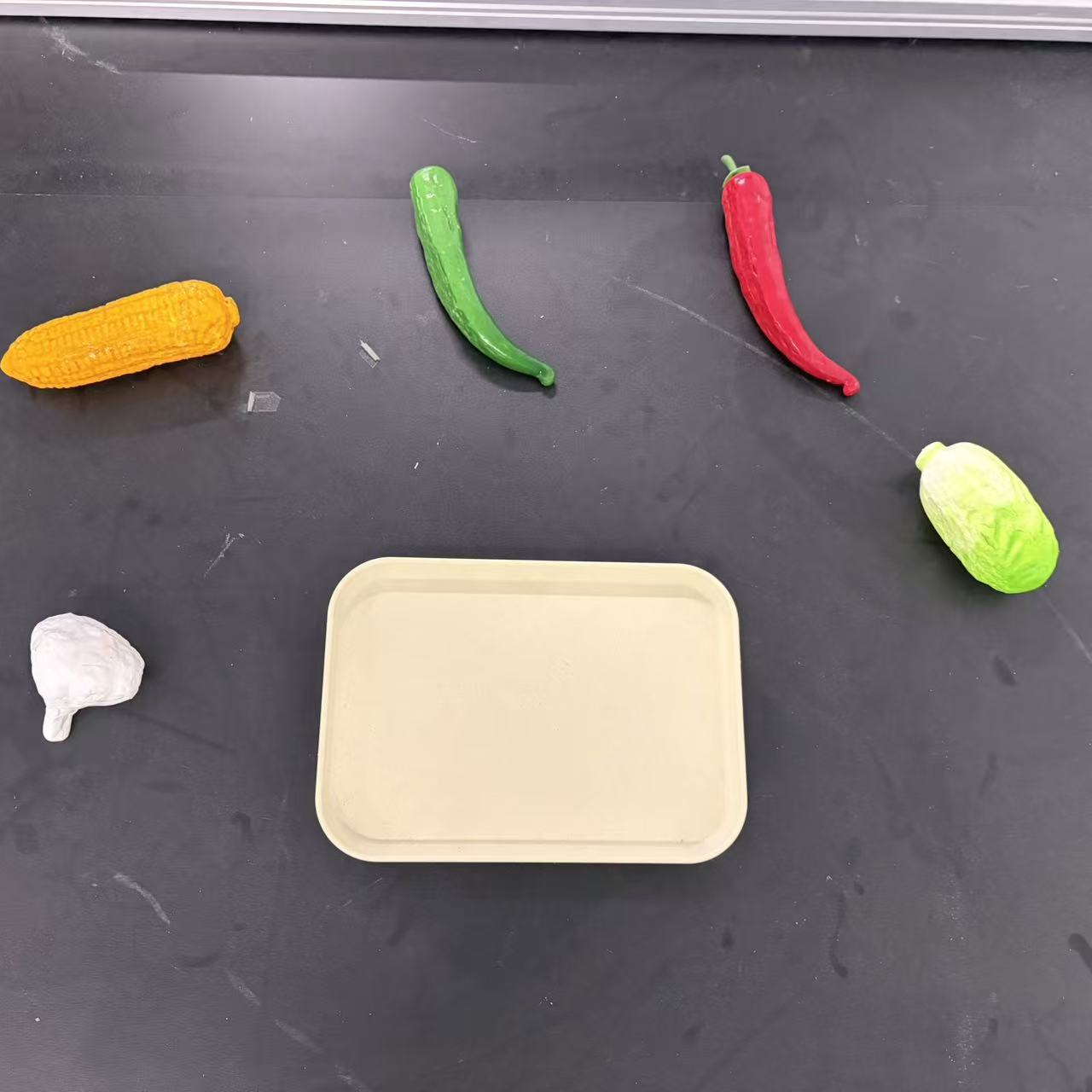}
    \vspace{2pt}
    \small Franka tabletop setup
\end{minipage}
\caption{
\textbf{In-lab \texttt{place\_object\_on\_plate} setup.}
The robot places one of five objects onto a fixed target plate.
}
\label{fig:app_inlab_scene}
\end{figure}

\paragraph{Franka \texttt{place\_object\_on\_plate}.}
The robot picks up one of five objects---corn, cabbage, green pepper, red pepper, and garlic---and places it onto a fixed plate (Fig.~\ref{fig:app_inlab_scene}).
A single fine-tuned checkpoint is shared across all five object variants.
A trial succeeds only if the object has been released and remains fully inside the plate at the end of the trial; otherwise it fails.
Successful trials receive 1 point and failures 0, so Score equals the success rate in percentage points.
Table~\ref{tab:app_place_object_on_plate_results} reports the per-object outcomes.

\begin{table}[h]
\centering
\small
\setlength{\tabcolsep}{6pt}
\renewcommand{\arraystretch}{1.12}
\caption{
\textbf{Per-object results for \texttt{place\_object\_on\_plate}.}
Each object variant is evaluated over 20 trials per method.
}
\label{tab:app_place_object_on_plate_results}
\begin{tabular}{@{}lccc@{}}
\toprule
Object & Baseline & PACE & \(\Delta\) \\
\midrule
Corn & 16 / 20 (80.0\%) & \best{18 / 20 (90.0\%)} & \gain{10.0} \\
Cabbage & 15 / 20 (75.0\%) & \best{17 / 20 (85.0\%)} & \gain{10.0} \\
Green pepper & 13 / 20 (65.0\%) & \best{18 / 20 (90.0\%)} & \gain{25.0} \\
Red pepper & 14 / 20 (70.0\%) & \best{17 / 20 (85.0\%)} & \gain{15.0} \\
Garlic & 14 / 20 (70.0\%) & \best{18 / 20 (90.0\%)} & \gain{20.0} \\
\midrule
\textbf{Total} & \textbf{72 / 100 (72.0\%)} & \best{88 / 100 (88.0\%)} & \gain{16.0} \\
\bottomrule
\end{tabular}
\end{table}

\subsection{Fine-tuning configuration}
\label{app:real_world_training}

For each RoboChallenge task, we fine-tune a task-specific \(\pi_{0.5}\) policy
initialized from the PI0.5 base checkpoint. Training is performed on four H200
GPUs with FSDP and bfloat16 precision. We use AdamW with a global batch size of
\(256\), global gradient clipping at norm \(1.0\), EMA decay \(0.99\), and eight
data-loader workers. The learning rate follows a cosine decay schedule with
\(1000\) warmup steps, peak learning rate \(2.5\times10^{-5}\), decay horizon
\(30{,}000\) steps, and final learning rate \(2.5\times10^{-6}\). For all
reported RoboChallenge trials, we deploy the \(30{,}000\)-step checkpoint.

For \texttt{place\_object\_on\_plate}, we fine-tune a single \(\pi_{0.5}\)
checkpoint over all five object variants using 301 demonstrations. The training
recipe follows the RoboChallenge setting except that we use a batch size of
\(32\) and train on one RTX A6000 Pro GPU. We also deploy the
\(30{,}000\)-step checkpoint for all in-lab trials.

\subsection{Evaluation and fixed-horizon comparisons}
\label{app:real_world_eval_protocol}
\label{app:franka_stronger_baselines}

On RoboChallenge, we compare PACE with full-chunk execution (\(H=50\)).
On Franka, we additionally evaluate fixed \(H=5\) and \(H=25\).
PACE selects the executed prefix using the default configuration in Appendix~\ref{app:adaptive_horizon_details}, without task-specific horizon tuning during evaluation.

Table~\ref{tab:franka_stronger_baselines} reports success rates and policy queries per successful episode on Franka.

\begin{table}[!htbp]
\centering
\small
\setlength{\tabcolsep}{8pt}
\renewcommand{\arraystretch}{1.15}
\caption{
\textbf{Comparison with fixed-horizon execution on Franka.}
Query counts are averaged over each strategy's successful episodes.
}
\label{tab:franka_stronger_baselines}
\begin{tabular}{lccc}
\toprule
Execution & \makecell{Successes /\\trials}
& \makecell{Fixed \(H\) /\\PACE mean \(H\)}
& \makecell{Queries /\\successful episode} \\
\midrule
Fixed \(H=5\) & 87 / 100 & 5 & 36.22 \\
Fixed \(H=25\) & 82 / 100 & 25 & 6.36 \\
Fixed \(H=50\) & 72 / 100 & 50 & 3.14 \\
\oursrow PACE & \best{88 / 100} & 28.64 & 5.57 \\
\bottomrule
\end{tabular}
\end{table}

PACE achieves a success rate comparable to fixed \(H=5\) (\(88\%\) versus \(87\%\)), while using \(84.6\%\) fewer policy queries per successful episode.
Relative to fixed \(H=25\), PACE improves success by 6 percentage points and reduces policy queries per successful episode by \(12.4\%\).
Fixed \(H=50\) requires the fewest queries per successful episode, but its success rate is lower at \(72\%\).

\section{Additional PACE rollout visualizations}
\label{app:additional_rollout_visualizations}

These additional RoboTwin2.0 and RoboChallenge rollouts show the policy-query timesteps and execution horizons selected by PACE.
Each caption summarizes how the horizon changes across the task's motion phases.

\subsection{RoboTwin2.0 rollouts}

\begin{figure}[H]
\centering
\includegraphics[width=\linewidth]{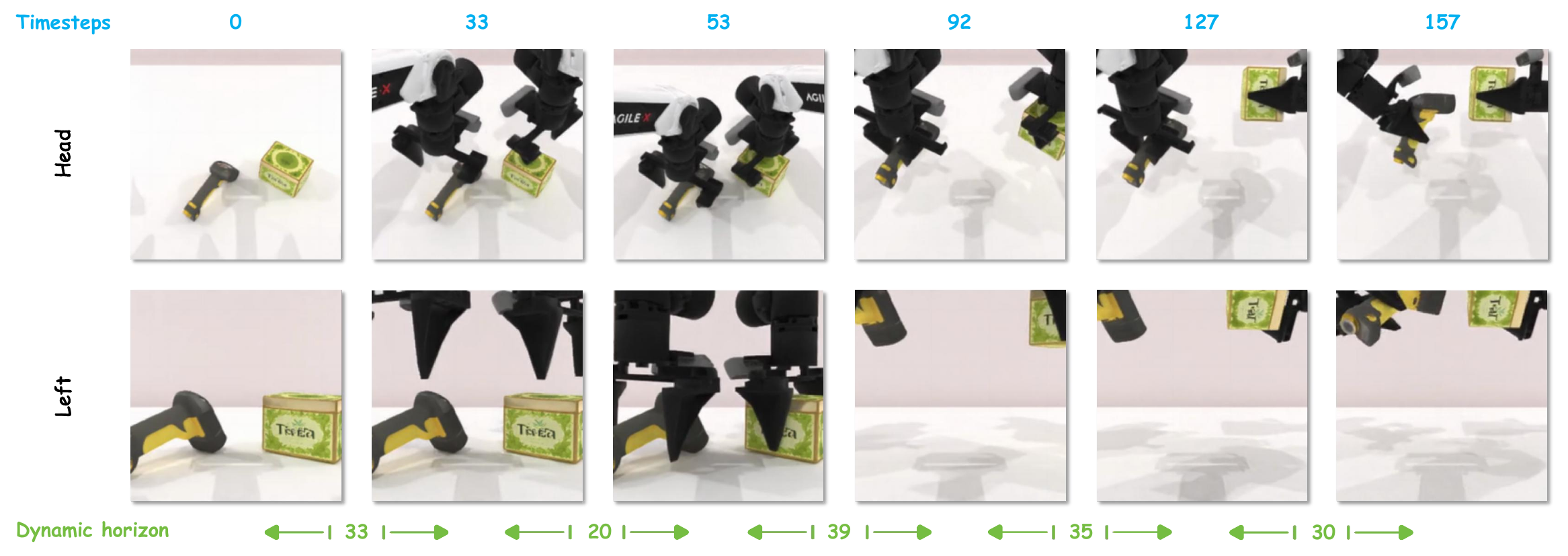}
\caption{
\textbf{PACE rollout visualization on \texttt{scan\_object}.}
PACE selects execution horizons from the predicted kinematic profile, allowing
the rollout to keep longer open-loop segments when motion is smooth and to
replan earlier around phase transitions.
}
\label{fig:app_scan_object_rollout}
\end{figure}

\clearpage
\begin{figure}[H]
\centering
\includegraphics[width=\linewidth]{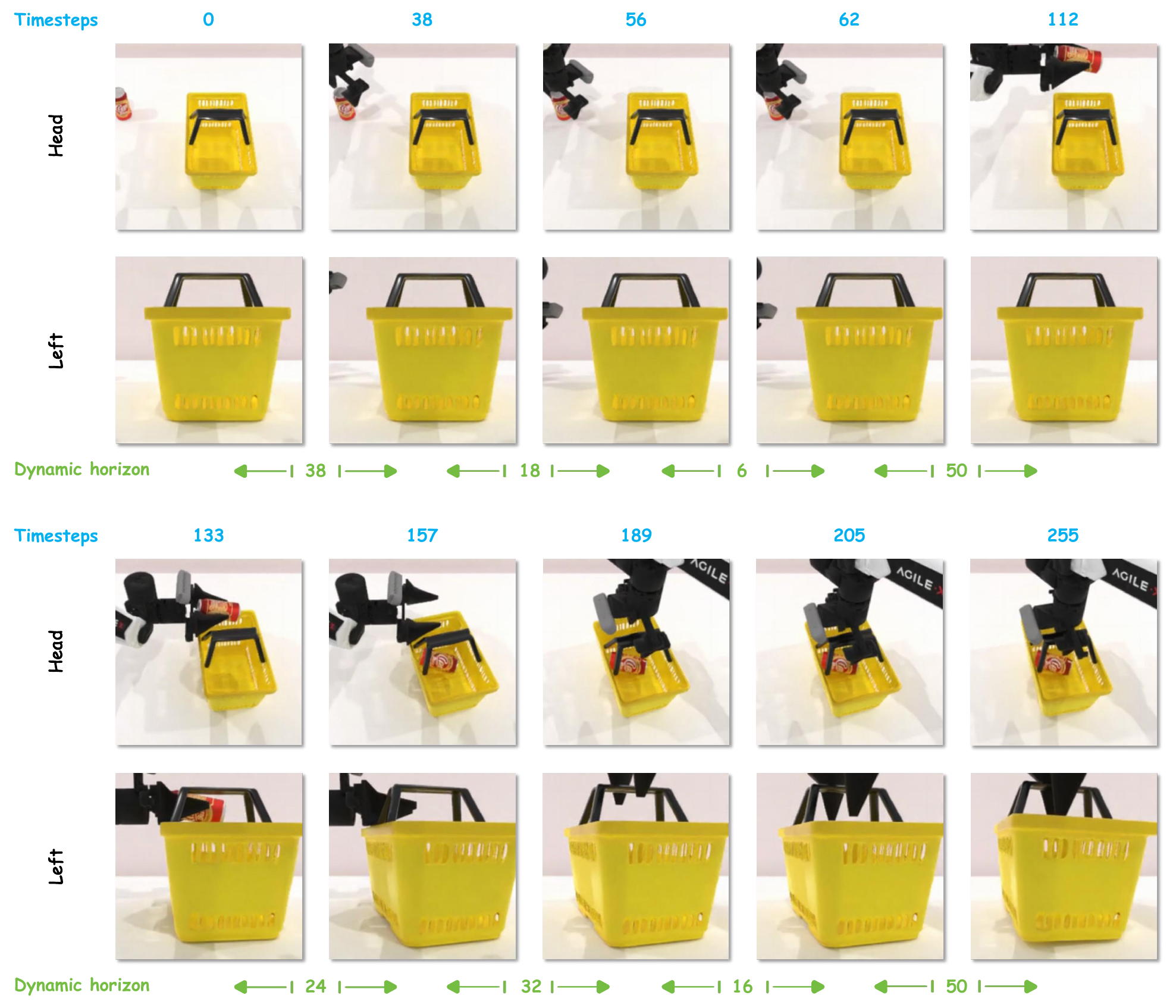}
\caption{
\textbf{PACE rollout visualization on \texttt{place\_can\_basket}.}
The selected execution horizon adapts to the local phase structure of the
rollout, with shorter prefixes near the placement transition and longer
prefixes during smooth motion segments.
}
\label{fig:app_place_can_basket_rollout}
\end{figure}

\clearpage
\begin{figure}[H]
\centering
\includegraphics[width=\linewidth]{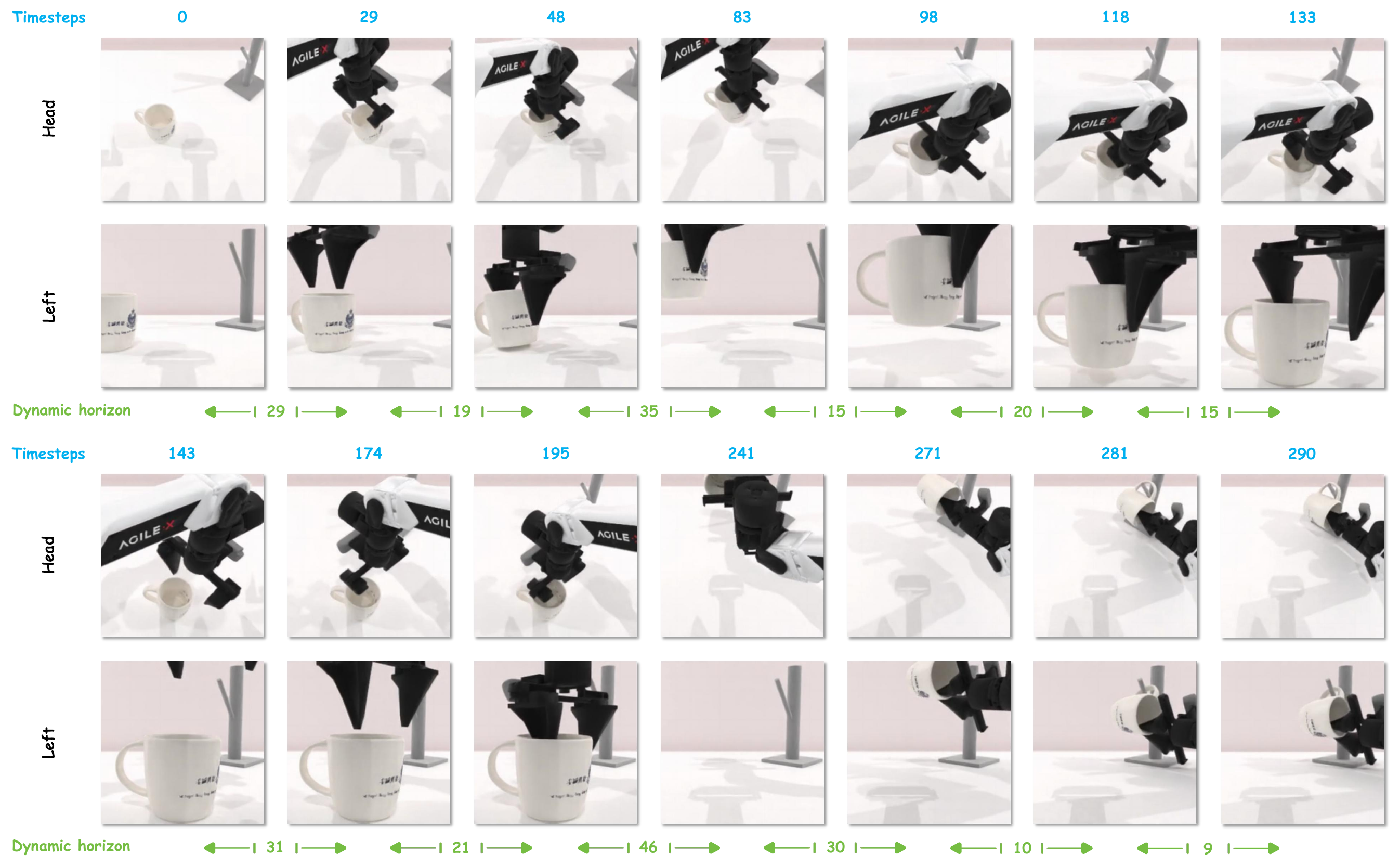}
\caption{
\textbf{PACE rollout visualization on \texttt{hanging\_mug}.}
The execution horizon is shortened around the task transition that requires
precise alignment and interaction, and lengthened during more stable motion
segments.
}
\label{fig:app_hanging_mug_rollout}
\end{figure}

\subsection{RoboChallenge rollout}

\begin{figure}[H]
\centering
\includegraphics[width=\linewidth]{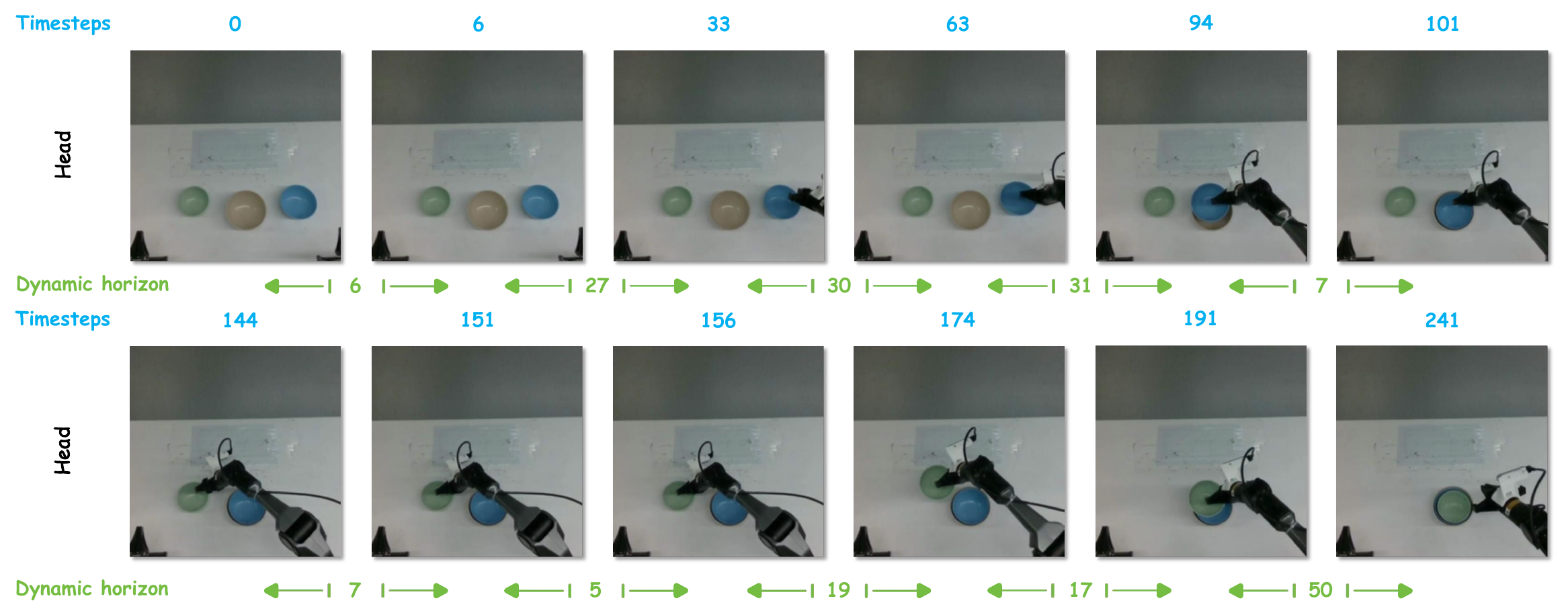}
\caption{
\textbf{PACE rollout visualization on RoboChallenge \texttt{stack\_bowls}.}
The selected execution horizon is longer during smooth approach and transport,
and shorter near the stacking phase, reflecting the same phase-aware replanning
rule used throughout the paper.
}
\label{fig:app_stack_bowls_rollout}
\end{figure}


\clearpage
\input{checklist.tex}

\end{document}

%% file: checklist.tex
\section*{NeurIPS Paper Checklist}

\begin{enumerate}

\item {\bf Claims}
    \item[] Question: Do the main claims made in the abstract and introduction accurately reflect the paper's contributions and scope?
    \item[] Answer: \answerYes{}.
    \item[] Justification: The abstract and Section~\ref{sec:intro} state the main claims: execution horizon is a critical test-time variable, fixed horizons are task-dependent and non-monotonic, and PACE improves simulation and real-robot execution without retraining. These claims are supported by the simulation results in Section~\ref{sec:main_results}, real-robot results in Section~\ref{sec:real_world_results}, and diagnostic analyses in Sections~\ref{sec:dynamic_window_visualization} and~\ref{sec:fixed_horizon_sweep}.
    \item[] Guidelines:
    \begin{itemize}
        \item The answer \answerNA{} means that the abstract and introduction do not include the claims made in the paper.
        \item The abstract and/or introduction should clearly state the claims made, including the contributions made in the paper and important assumptions and limitations. A \answerNo{} or \answerNA{} answer to this question will not be perceived well by the reviewers. 
        \item The claims made should match theoretical and experimental results, and reflect how much the results can be expected to generalize to other settings. 
        \item It is fine to include aspirational goals as motivation as long as it is clear that these goals are not attained by the paper. 
    \end{itemize}

\item {\bf Limitations}
    \item[] Question: Does the paper discuss the limitations of the work performed by the authors?
    \item[] Answer: \answerYes{}.
    \item[] Justification: Sec.~\ref{sec:limitations} discusses PACE's dependence on the base policy, reliance on low-speed valley cues, and need for demonstration-based calibration.
It also addresses the scope of the empirical evaluation and the inference cost of additional policy queries.
    \item[] Guidelines:
    \begin{itemize}
        \item The answer \answerNA{} means that the paper has no limitation while the answer \answerNo{} means that the paper has limitations, but those are not discussed in the paper. 
        \item The authors are encouraged to create a separate ``Limitations'' section in their paper.
        \item The paper should point out any strong assumptions and how robust the results are to violations of these assumptions (e.g., independence assumptions, noiseless settings, model well-specification, asymptotic approximations only holding locally). The authors should reflect on how these assumptions might be violated in practice and what the implications would be.
        \item The authors should reflect on the scope of the claims made, e.g., if the approach was only tested on a few datasets or with a few runs. In general, empirical results often depend on implicit assumptions, which should be articulated.
        \item The authors should reflect on the factors that influence the performance of the approach. For example, a facial recognition algorithm may perform poorly when image resolution is low or images are taken in low lighting. Or a speech-to-text system might not be used reliably to provide closed captions for online lectures because it fails to handle technical jargon.
        \item The authors should discuss the computational efficiency of the proposed algorithms and how they scale with dataset size.
        \item If applicable, the authors should discuss possible limitations of their approach to address problems of privacy and fairness.
        \item While the authors might fear that complete honesty about limitations might be used by reviewers as grounds for rejection, a worse outcome might be that reviewers discover limitations that aren't acknowledged in the paper. The authors should use their best judgment and recognize that individual actions in favor of transparency play an important role in developing norms that preserve the integrity of the community. Reviewers will be specifically instructed to not penalize honesty concerning limitations.
    \end{itemize}

\item {\bf Theory assumptions and proofs}
    \item[] Question: For each theoretical result, does the paper provide the full set of assumptions and a complete (and correct) proof?
    \item[] Answer: \answerNA{}.
    \item[] Justification: The paper does not present theoretical results or formal proofs. The equations in Sections~\ref{sec:method} and~\ref{app:adaptive_horizon_details} define the action-chunking setup, speed profiles, and deterministic PACE selection rule.
    \item[] Guidelines:
    \begin{itemize}
        \item The answer \answerNA{} means that the paper does not include theoretical results. 
        \item All the theorems, formulas, and proofs in the paper should be numbered and cross-referenced.
        \item All assumptions should be clearly stated or referenced in the statement of any theorems.
        \item The proofs can either appear in the main paper or the supplemental material, but if they appear in the supplemental material, the authors are encouraged to provide a short proof sketch to provide intuition. 
        \item Inversely, any informal proof provided in the core of the paper should be complemented by formal proofs provided in appendix or supplemental material.
        \item Theorems and Lemmas that the proof relies upon should be properly referenced. 
    \end{itemize}

\item {\bf Experimental result reproducibility}
    \item[] Question: Does the paper fully disclose all the information needed to reproduce the main experimental results of the paper to the extent that it affects the main claims and/or conclusions of the paper (regardless of whether the code and data are provided or not)?
    \item[] Answer: \answerYes{}.
    \item[] Justification: Sections~\ref{sec:method} and~\ref{app:adaptive_horizon_details} specify the PACE algorithm, speed signal, smoothing, threshold calibration, and multi-arm fusion rule. Sections~\ref{sec:sim_experiments},~\ref{app:sim_full_results},~\ref{app:ablation_details},~\ref{app:fixed_horizon_sweep}, and~\ref{app:real_world_details} describe the benchmarks, seeds, trial counts, baselines, training settings, and evaluation protocols.
    \item[] Guidelines:
    \begin{itemize}
        \item The answer \answerNA{} means that the paper does not include experiments.
        \item If the paper includes experiments, a \answerNo{} answer to this question will not be perceived well by the reviewers: Making the paper reproducible is important, regardless of whether the code and data are provided or not.
        \item If the contribution is a dataset and\slash or model, the authors should describe the steps taken to make their results reproducible or verifiable. 
        \item Depending on the contribution, reproducibility can be accomplished in various ways. For example, if the contribution is a novel architecture, describing the architecture fully might suffice, or if the contribution is a specific model and empirical evaluation, it may be necessary to either make it possible for others to replicate the model with the same dataset, or provide access to the model. In general. releasing code and data is often one good way to accomplish this, but reproducibility can also be provided via detailed instructions for how to replicate the results, access to a hosted model (e.g., in the case of a large language model), releasing of a model checkpoint, or other means that are appropriate to the research performed.
        \item While NeurIPS does not require releasing code, the conference does require all submissions to provide some reasonable avenue for reproducibility, which may depend on the nature of the contribution. For example
        \begin{enumerate}
            \item If the contribution is primarily a new algorithm, the paper should make it clear how to reproduce that algorithm.
            \item If the contribution is primarily a new model architecture, the paper should describe the architecture clearly and fully.
            \item If the contribution is a new model (e.g., a large language model), then there should either be a way to access this model for reproducing the results or a way to reproduce the model (e.g., with an open-source dataset or instructions for how to construct the dataset).
            \item We recognize that reproducibility may be tricky in some cases, in which case authors are welcome to describe the particular way they provide for reproducibility. In the case of closed-source models, it may be that access to the model is limited in some way (e.g., to registered users), but it should be possible for other researchers to have some path to reproducing or verifying the results.
        \end{enumerate}
    \end{itemize}

\item {\bf Open access to data and code}
    \item[] Question: Does the paper provide open access to the data and code, with sufficient instructions to faithfully reproduce the main experimental results, as described in supplemental material?
    \item[] Answer: \answerNo{}.
    \item[] Justification: Code and usage instructions are available through the \href{https://niejunnan25.github.io/pace-project/}{PACE project page}.
The release covers a subset of the reported experiments.
    \item[] Guidelines:
    \begin{itemize}
        \item The answer \answerNA{} means that paper does not include experiments requiring code.
        \item Please see the NeurIPS code and data submission guidelines (\url{https://neurips.cc/public/guides/CodeSubmissionPolicy}) for more details.
        \item While we encourage the release of code and data, we understand that this might not be possible, so \answerNo{} is an acceptable answer. Papers cannot be rejected simply for not including code, unless this is central to the contribution (e.g., for a new open-source benchmark).
        \item The instructions should contain the exact command and environment needed to run to reproduce the results. See the NeurIPS code and data submission guidelines (\url{https://neurips.cc/public/guides/CodeSubmissionPolicy}) for more details.
        \item The authors should provide instructions on data access and preparation, including how to access the raw data, preprocessed data, intermediate data, and generated data, etc.
        \item The authors should provide scripts to reproduce all experimental results for the new proposed method and baselines. If only a subset of experiments are reproducible, they should state which ones are omitted from the script and why.
        \item At submission time, to preserve anonymity, the authors should release anonymized versions (if applicable).
        \item Providing as much information as possible in supplemental material (appended to the paper) is recommended, but including URLs to data and code is permitted.
    \end{itemize}

\item {\bf Experimental setting/details}
    \item[] Question: Does the paper specify all the training and test details (e.g., data splits, hyperparameters, how they were chosen, type of optimizer) necessary to understand the results?
    \item[] Answer: \answerYes{}.
    \item[] Justification: The main text and appendix specify the prediction horizon, execution baselines, PACE configuration, threshold calibration from demonstrations, seeds, episode counts, ablation settings, and real-robot trial counts. Appendix~\ref{app:real_world_training} reports optimizer, batch size, learning-rate schedule, precision, checkpoint step, and hardware for real-robot fine-tuning.
    \item[] Guidelines:
    \begin{itemize}
        \item The answer \answerNA{} means that the paper does not include experiments.
        \item The experimental setting should be presented in the core of the paper to a level of detail that is necessary to appreciate the results and make sense of them.
        \item The full details can be provided either with the code, in appendix, or as supplemental material.
    \end{itemize}

\item {\bf Experiment statistical significance}
    \item[] Question: Does the paper report error bars suitably and correctly defined or other appropriate information about the statistical significance of the experiments?
    \item[] Answer: \answerNo{}.
    \item[] Justification: The simulation results and phase-alignment analysis include uncertainty estimates (Appendices~\ref{app:adaptive_baseline_statistics}, \ref{app:libero}, \ref{app:shared_threshold}, and~\ref{app:phase_alignment}).
The main real-robot results are reported without uncertainty estimates.
    \item[] Guidelines:
    \begin{itemize}
        \item The answer \answerNA{} means that the paper does not include experiments.
        \item The authors should answer \answerYes{} if the results are accompanied by error bars, confidence intervals, or statistical significance tests, at least for the experiments that support the main claims of the paper.
        \item The factors of variability that the error bars are capturing should be clearly stated (for example, train/test split, initialization, random drawing of some parameter, or overall run with given experimental conditions).
        \item The method for calculating the error bars should be explained (closed form formula, call to a library function, bootstrap, etc.)
        \item The assumptions made should be given (e.g., Normally distributed errors).
        \item It should be clear whether the error bar is the standard deviation or the standard error of the mean.
        \item It is OK to report 1-sigma error bars, but one should state it. The authors should preferably report a 2-sigma error bar than state that they have a 96\% CI, if the hypothesis of Normality of errors is not verified.
        \item For asymmetric distributions, the authors should be careful not to show in tables or figures symmetric error bars that would yield results that are out of range (e.g., negative error rates).
        \item If error bars are reported in tables or plots, the authors should explain in the text how they were calculated and reference the corresponding figures or tables in the text.
    \end{itemize}

\item {\bf Experiments compute resources}
    \item[] Question: For each experiment, does the paper provide sufficient information on the computer resources (type of compute workers, memory, time of execution) needed to reproduce the experiments?
    \item[] Answer: \answerNo{}.
    \item[] Justification: Appendix~\ref{app:real_world_training} reports the GPU setup used for real-robot fine-tuning, and the simulation and test-time execution experiments were run on internal GPU servers with two Intel Xeon Platinum 8468V sockets, approximately 1.9 TiB system memory, and eight NVIDIA H20 GPUs with 97{,}871 MiB memory each. However, the current manuscript does not provide precise wall-clock runtimes, GPU-hours, or total compute budget for each experiment.
    \item[] Guidelines:
    \begin{itemize}
        \item The answer \answerNA{} means that the paper does not include experiments.
        \item The paper should indicate the type of compute workers CPU or GPU, internal cluster, or cloud provider, including relevant memory and storage.
        \item The paper should provide the amount of compute required for each of the individual experimental runs as well as estimate the total compute. 
        \item The paper should disclose whether the full research project required more compute than the experiments reported in the paper (e.g., preliminary or failed experiments that didn't make it into the paper). 
    \end{itemize}
    
\item {\bf Code of ethics}
    \item[] Question: Does the research conducted in the paper conform, in every respect, with the NeurIPS Code of Ethics \url{https://neurips.cc/public/EthicsGuidelines}?
    \item[] Answer: \answerYes{}.
    \item[] Justification: The work uses simulation benchmarks and controlled robot manipulation experiments, and does not involve private data, human-subject studies, or deceptive user-facing deployment. We have reviewed the NeurIPS Code of Ethics and are not aware of any deviation.
    \item[] Guidelines:
    \begin{itemize}
        \item The answer \answerNA{} means that the authors have not reviewed the NeurIPS Code of Ethics.
        \item If the authors answer \answerNo, they should explain the special circumstances that require a deviation from the Code of Ethics.
        \item The authors should make sure to preserve anonymity (e.g., if there is a special consideration due to laws or regulations in their jurisdiction).
    \end{itemize}

\item {\bf Broader impacts}
    \item[] Question: Does the paper discuss both potential positive societal impacts and negative societal impacts of the work performed?
    \item[] Answer: \answerNo{}.
    \item[] Justification: The current manuscript focuses on test-time execution for robot policies and does not include a dedicated broader-impact discussion. Potential positive impacts include more reliable robot execution without retraining, while negative impacts would mainly arise from deploying more capable robot policies without appropriate real-world safety controls.
    \item[] Guidelines:
    \begin{itemize}
        \item The answer \answerNA{} means that there is no societal impact of the work performed.
        \item If the authors answer \answerNA{} or \answerNo, they should explain why their work has no societal impact or why the paper does not address societal impact.
        \item Examples of negative societal impacts include potential malicious or unintended uses (e.g., disinformation, generating fake profiles, surveillance), fairness considerations (e.g., deployment of technologies that could make decisions that unfairly impact specific groups), privacy considerations, and security considerations.
        \item The conference expects that many papers will be foundational research and not tied to particular applications, let alone deployments. However, if there is a direct path to any negative applications, the authors should point it out. For example, it is legitimate to point out that an improvement in the quality of generative models could be used to generate Deepfakes for disinformation. On the other hand, it is not needed to point out that a generic algorithm for optimizing neural networks could enable people to train models that generate Deepfakes faster.
        \item The authors should consider possible harms that could arise when the technology is being used as intended and functioning correctly, harms that could arise when the technology is being used as intended but gives incorrect results, and harms following from (intentional or unintentional) misuse of the technology.
        \item If there are negative societal impacts, the authors could also discuss possible mitigation strategies (e.g., gated release of models, providing defenses in addition to attacks, mechanisms for monitoring misuse, mechanisms to monitor how a system learns from feedback over time, improving the efficiency and accessibility of ML).
    \end{itemize}
    
\item {\bf Safeguards}
    \item[] Question: Does the paper describe safeguards that have been put in place for responsible release of data or models that have a high risk for misuse (e.g., pre-trained language models, image generators, or scraped datasets)?
    \item[] Answer: \answerNA{}.
    \item[] Justification: We release PACE code and task-specific RoboTwin policy checkpoints for manipulation research.
The release does not include high-risk models or datasets.
    \item[] Guidelines:
    \begin{itemize}
        \item The answer \answerNA{} means that the paper poses no such risks.
        \item Released models that have a high risk for misuse or dual-use should be released with necessary safeguards to allow for controlled use of the model, for example by requiring that users adhere to usage guidelines or restrictions to access the model or implementing safety filters. 
        \item Datasets that have been scraped from the Internet could pose safety risks. The authors should describe how they avoided releasing unsafe images.
        \item We recognize that providing effective safeguards is challenging, and many papers do not require this, but we encourage authors to take this into account and make a best faith effort.
    \end{itemize}

\item {\bf Licenses for existing assets}
    \item[] Question: Are the creators or original owners of assets (e.g., code, data, models), used in the paper, properly credited and are the license and terms of use explicitly mentioned and properly respected?
    \item[] Answer: \answerNo{}.
    \item[] Justification: The paper cites the original sources. \href{https://github.com/Physical-Intelligence/openpi/blob/main/LICENSE}{OpenPI}, \href{https://github.com/huggingface/lerobot/blob/main/LICENSE}{LeRobot}, \href{https://github.com/NVIDIA/Isaac-GR00T/blob/main/LICENSE}{GR00T}, and \href{https://github.com/hatchetProject/AutoHorizon/blob/c7504f1756109103f2cfcc2e23f1b1a23841c885/LICENSE}{AutoHorizon} code use Apache License 2.0. \href{https://github.com/Adaptive-Action-Chunking/libero/blob/ff8c8e421f7805b5845a72d07fbd8f1171df74bf/LICENSE}{AAC's LIBERO implementation}, \href{https://github.com/Lifelong-Robot-Learning/LIBERO/blob/master/LICENSE}{LIBERO code}, and \href{https://huggingface.co/datasets/TianxingChen/RoboTwin2.0}{RoboTwin2.0 code/data} use MIT. The PaliGemma backbone of \(\pi_{0.5}\) is subject to the \href{https://ai.google.dev/gemma/terms}{Gemma Terms of Use}. \href{https://huggingface.co/nvidia/GR00T-N1.7-LIBERO}{GR00T N1.7 weights} use the \href{https://www.nvidia.com/en-us/agreements/enterprise-software/nvidia-open-model-license/}{NVIDIA Open Model License Agreement}, distinct from the code license. \href{https://github.com/Lifelong-Robot-Learning/LIBERO#license}{LIBERO demonstrations} and \href{https://arxiv.org/html/2403.12945v2#S3}{DROID data} use CC BY 4.0. We access DROID through the \href{https://huggingface.co/datasets/GEAR-Dreams/DreamZero-DROID-Data}{DreamZero conversion}. \href{https://huggingface.co/datasets/x-humanoid-robomind/RoboMIND}{RoboMIND 1.x}, \href{https://modelscope.cn/datasets/X-Humanoid/RoboMIND2.0-Franka-Part-1}{RoboMIND 2.0}, and \href{https://github.com/FlagOpen/RoboCOIN}{RoboCOIN} data use Apache License 2.0. \href{https://huggingface.co/datasets/agibot-world/AgiBotWorld-Beta}{AgiBot World Beta data} and \href{https://github.com/RoboChallenge/RoboChallengeInference/blob/main/LICENSE}{RoboChallenge inference code} use CC BY-NC-SA 4.0, with attribution and noncommercial-use conditions, and share-alike conditions when adaptations are shared. The consulted \href{https://huggingface.co/datasets/RoboChallenge/Table30}{Table30 V1 demonstration releases} do not specify a data license.
    \item[] Guidelines:
    \begin{itemize}
        \item The answer \answerNA{} means that the paper does not use existing assets.
        \item The authors should cite the original paper that produced the code package or dataset.
        \item The authors should state which version of the asset is used and, if possible, include a URL.
        \item The name of the license (e.g., CC-BY 4.0) should be included for each asset.
        \item For scraped data from a particular source (e.g., website), the copyright and terms of service of that source should be provided.
        \item If assets are released, the license, copyright information, and terms of use in the package should be provided. For popular datasets, \url{paperswithcode.com/datasets} has curated licenses for some datasets. Their licensing guide can help determine the license of a dataset.
        \item For existing datasets that are re-packaged, both the original license and the license of the derived asset (if it has changed) should be provided.
        \item If this information is not available online, the authors are encouraged to reach out to the asset's creators.
    \end{itemize}

\item {\bf New assets}
    \item[] Question: Are new assets introduced in the paper well documented and is the documentation provided alongside the assets?
    \item[] Answer: \answerYes{}.
    \item[] Justification: The released code includes usage and licensing documentation.
The \href{https://huggingface.co/niejunnan25/pi05-robotwin}{RoboTwin policy checkpoints} are accompanied by a model card describing training and usage, together with the applicable model terms.
    \item[] Guidelines:
    \begin{itemize}
        \item The answer \answerNA{} means that the paper does not release new assets.
        \item Researchers should communicate the details of the dataset\slash code\slash model as part of their submissions via structured templates. This includes details about training, license, limitations, etc. 
        \item The paper should discuss whether and how consent was obtained from people whose asset is used.
        \item At submission time, remember to anonymize your assets (if applicable). You can either create an anonymized URL or include an anonymized zip file.
    \end{itemize}

\item {\bf Crowdsourcing and research with human subjects}
    \item[] Question: For crowdsourcing experiments and research with human subjects, does the paper include the full text of instructions given to participants and screenshots, if applicable, as well as details about compensation (if any)? 
    \item[] Answer: \answerNA{}.
    \item[] Justification: The paper does not involve crowdsourcing, user studies, or research with human subjects.
    \item[] Guidelines:
    \begin{itemize}
        \item The answer \answerNA{} means that the paper does not involve crowdsourcing nor research with human subjects.
        \item Including this information in the supplemental material is fine, but if the main contribution of the paper involves human subjects, then as much detail as possible should be included in the main paper. 
        \item According to the NeurIPS Code of Ethics, workers involved in data collection, curation, or other labor should be paid at least the minimum wage in the country of the data collector. 
    \end{itemize}

\item {\bf Institutional review board (IRB) approvals or equivalent for research with human subjects}
    \item[] Question: Does the paper describe potential risks incurred by study participants, whether such risks were disclosed to the subjects, and whether Institutional Review Board (IRB) approvals (or an equivalent approval/review based on the requirements of your country or institution) were obtained?
    \item[] Answer: \answerNA{}.
    \item[] Justification: The paper does not involve crowdsourcing, user studies, or research with human subjects, so IRB approval is not applicable.
    \item[] Guidelines:
    \begin{itemize}
        \item The answer \answerNA{} means that the paper does not involve crowdsourcing nor research with human subjects.
        \item Depending on the country in which research is conducted, IRB approval (or equivalent) may be required for any human subjects research. If you obtained IRB approval, you should clearly state this in the paper. 
        \item We recognize that the procedures for this may vary significantly between institutions and locations, and we expect authors to adhere to the NeurIPS Code of Ethics and the guidelines for their institution. 
        \item For initial submissions, do not include any information that would break anonymity (if applicable), such as the institution conducting the review.
    \end{itemize}

\begin{samepage}
\item {\bf Declaration of LLM usage}
    \item[] Question: Does the paper describe the usage of LLMs if it is an important, original, or non-standard component of the core methods in this research? Note that if the LLM is used only for writing, editing, or formatting purposes and does \emph{not} impact the core methodology, scientific rigor, or originality of the research, declaration is not required.
    \item[] Answer: \answerYes{}.
    \item[] Justification: The paper studies vision-language-action robot policies and describes the use of language-conditioned robot policies such as \(\pi_{0.5}\) throughout Sections~\ref{sec:intro},~\ref{sec:related},~\ref{sec:method}, and~\ref{sec:real_world}. LLMs used only for writing, editing, or formatting are not material to the research claims.
    \item[] Guidelines:
    \begin{itemize}
        \item The answer \answerNA{} means that the core method development in this research does not involve LLMs as any important, original, or non-standard components.
        \item Please refer to our LLM policy in the NeurIPS handbook for what should or should not be described.
    \end{itemize}
\end{samepage}

\end{enumerate}